%% file: survey.tex
\documentclass[lettersize,journal]{IEEEtran}
\usepackage{amsmath,amsfonts}
\usepackage{algorithmic}
\usepackage{algorithm}
\usepackage{array}
\usepackage[caption=false,font=normalsize,labelfont=sf,textfont=sf]{subfig}
\usepackage{textcomp}
\usepackage{stfloats}
\usepackage{url}
\usepackage{verbatim}
\usepackage{graphicx}
\usepackage{cite}
\usepackage{multirow}
\usepackage{multicol}
\usepackage{booktabs}
\usepackage{bm}
\usepackage{wrapfig}
\usepackage[table,dvipsnames]{xcolor}
\usepackage{pifont}
\usepackage{makecell}

\usepackage{tikz}
\usepackage[edges]{forest}
\definecolor{rootcolor}{RGB}{0, 104, 139}
\definecolor{level1color}{RGB}{72, 145, 173}
\definecolor{level2color}{RGB}{144, 186, 201}
\definecolor{level3color}{RGB}{216, 227, 231}
\definecolor{textcolor}{RGB}{51, 51, 51}
\definecolor{mydarkblue}{rgb}{0,0.08,0.45}

\def\etal{\emph{et al.}}
\def\eg{\emph{e.g.}}

\newcommand{\mcite}[2]{#1~\etal~\cite{#2}}

\newcommand{\cmark}{\ding{51}}%
\newcommand{\xmark}{\ding{55}}%

\usepackage[pagebackref=true,breaklinks=true,letterpaper=true,colorlinks,bookmarks=false]{hyperref}

\hypersetup{
citecolor=rootcolor,
}

\begin{document}

\title{Low-Precision Training of Large Language Models: Methods, Challenges, and Opportunities}

\author{Zhiwei~Hao, Jianyuan~Guo, Li~Shen, Yong~Luo, Han~Hu, Guoxia~Wang, Dianhai~Yu,\\Yonggang~Wen, Dacheng~Tao
\thanks{Zhiwei~Hao and Han~Hu are with the School of Information and Electronics, Beijing Institute of Technology, Beijing 100081, China Email: haozhw@outlook.com; hhu@bit.edu.cn}
\thanks{Jianyuan Guo is with the Department of Computer Science, City University of Hong Kong, Hong Kong 999077, China. Email: jianyguo@cityu.edu.hk}
\thanks{Li Shen is with the School of Cyber Science and Technology, Shenzhen Campus of Sun Yat-sen University, Shenzhen 518107, China. Email: shenli6@mail.sysu.edu.cn}
\thanks{Yong~Luo is with the School of Computer Science, Wuhan University, Wuhan 430072, China. Email: luoyong@whu.edu.cn}
\thanks{Guoxia Wang and Dianhai Yu are with the Baidu Inc., Beijing 100000, China. Email: wangguoxia@baidu.com; yudianhai@baidu.com}
\thanks{Yonggang~Wen and Dacheng~Tao are with the College of Computing and Data Science, Nanyang Technological University, 639798, Singapore. Email: ygwen@ntu.edu.sg; dacheng.tao@gmail.com}
}

\maketitle

\begin{abstract}
    Large language models (LLMs) have achieved impressive performance across various domains. However, the substantial hardware resources required for their training present a significant barrier to efficiency and scalability. To mitigate this challenge, low-precision training techniques have been widely adopted, leading to notable advancements in training efficiency. Despite these gains, low-precision training involves several components, such as weights, activations, and gradients, each of which can be represented in different numerical formats. The resulting diversity has created a fragmented landscape in low-precision training research, making it difficult for researchers to gain a unified overview of the field.
    This survey provides a comprehensive review of existing low-precision training methods. To systematically organize these approaches, we categorize them into three primary groups based on their underlying numerical formats, which is a key factor influencing hardware compatibility, computational efficiency, and ease of reference for readers. The categories are (1) fixed-point and integer-based methods, (2) floating-point-based methods, and (3) customized format-based methods.
    Additionally, we discuss quantization-aware training approaches, which share key similarities with low-precision training during forward propagation. 
    Beyond efficiency, we examine robustness and deployment reliability under low precision. Finally, we highlight several promising research directions to advance this field. 
    A collection of papers discussed in this survey is provided in \href{https://github.com/Hao840/Awesome-Low-Precision-Training}{Awesome-Low-Precision-Training}.
\end{abstract}

\begin{IEEEkeywords}
  large language models; low-precision training; quantization
\end{IEEEkeywords}

\section{Introduction}

\IEEEPARstart{L}{arge} Language Models (LLMs) have emerged as a foundational technology in modern artificial intelligence, driving breakthroughs in natural language processing, code generation, and multimodal reasoning~\cite{achiam2023gpt,team2024gemini,liu2024deepseek,yang2024qwen2}. Their capacity to model complex patterns across massive datasets has enabled a diverse array of applications, ranging from conversational agents to tools that accelerate scientific discovery. Despite their transformative potential, training LLMs remains prohibitively expensive, demanding extensive computational resources and incurring significant energy costs~\cite{shen2024efficient,strubell2019energy,patterson2021carbon,cottier2024rising}. For instance, GPT-3 required roughly 355 V100 GPU-years of training compute, and recent analyses indicate that the cost of frontier-model training has continued to rise rapidly~\cite{brown2020language,cottier2024rising}.

To address these challenges, low-precision training has emerged as a promising solution. By reducing the numerical precision of weights, gradients, and activations during training, such as switching from 32-bit floating-point (FP32) precision to 16-bit (FP16/BF16) or even 8-bit formats, researchers can significantly lower memory usage, communication overhead, and computational costs, while maintaining competitive model performance~\cite{micikeviciusnad18,kalamkar2019study,micikevicius2022fp8,peng2023fp8,liu2024deepseek}.

Low-precision training is attractive for large or resource-constrained models because it reduces memory use, data movement, and communication volume across parameters, activations, gradients, optimizer states, and exchanged tensors. For example, switching from FP32 to FP16 halves storage for the affected tensors, while low-bit optimizers further shrink stateful training overhead~\cite{micikeviciusnad18,dettmers20228bit,peng2023fp8}. Modern accelerators also expose faster execution paths for FP16/BF16, FP8, and related formats, and distributed recipes use gradient quantization, sign compression, and error feedback to reduce synchronization cost~\cite{kalamkar2019study,micikevicius2022fp8,luo2024ascend,seidefdly14,alistarhg0tv17,karimireddyrsj19}. These advantages make low precision central to scalable and resource-conscious LLM training.

In parallel, GPU architectures have evolved to better support low-precision computation. Table~\ref{tab:gpu} uses NVIDIA GPUs as one representative example. Hardware acceleration has rapidly shifted from the FP32-only Pascal generation to half-precision in Volta and Ampere, and recently to ultra-low precisions with native FP8, FP6, and FP4 support in Hopper and Blackwell Tensor Cores. Low-precision training support is not limited to CUDA-based systems. AMD Instinct accelerators and the ROCm software stack have become increasingly relevant for large-scale training infrastructure. For example, CDNA 3-based MI300X-class accelerators expose high-throughput BF16, FP16, INT8, and FP8 matrix paths, with ROCm support for native FP8 FNUZ E4M3/E5M2 on CDNA 3 hardware.
Similar trends toward richer low-precision support appear in other non-NVIDIA ecosystems, such as TPU-oriented TensorFlow/JAX stacks and Ascend-oriented formats including HiFloat8.
Recent MX data types such as MXFP8/MXFP4 further extend the design space through tighter hardware-software co-design~\cite{tensorflow2015,jax2018github,kalamkar2019study,micikevicius2022fp8,luo2024ascend,rouhani2023microscaling,tseng2025training}.

\begin{table}[t]
	\renewcommand\tabcolsep{5.0pt}
    \centering
    \caption{Comparison of supported precisions across NVIDIA GPUs.}
    \begin{tabular}{c|c|ccccc}
        \toprule
        \multicolumn{2}{c|}{GPU} & P4 & V100 & A100 & H100 & B100 \\
        \midrule
        \multicolumn{2}{c|}{Architecture} & Pascal & Volta & Ampere & Hopper & Blackwell \\
        \midrule
        \multirow{4}{*}{\makecell{CUDA\\Core}} & FP32 & \cmark & \cmark & \cmark & \cmark & \cmark \\
        & FP16 & \xmark & \cmark & \cmark & \cmark & \cmark \\
        & BF16 & \xmark & \xmark & \cmark & \cmark & \cmark \\
        & FP8 & \xmark & \xmark & \xmark & \xmark & \xmark \\
        & FP6/FP4 & \xmark & \xmark & \xmark & \xmark & \xmark \\
        \midrule
        \multirow{4}{*}{\makecell{Tensor\\Core}} & FP32 & \xmark & \xmark & \xmark & \xmark & \xmark \\
        & FP16 & \xmark & \cmark & \cmark & \cmark & \cmark \\
        & BF16 & \xmark & \xmark & \cmark & \cmark & \cmark \\
        & FP8 & \xmark & \xmark & \xmark & \cmark & \cmark \\
        & FP6/FP4 & \xmark & \xmark & \xmark & \xmark & \cmark \\
        \bottomrule
    \end{tabular}
    \label{tab:gpu}
\end{table}

Despite its promise, low-precision training remains difficult to survey because methods are fragmented across value representations, targeted training components, and training scenarios. Recent surveys provide valuable context on distributed LLM infrastructure or low-bit LLM quantization~\cite{duan2024efficient,gong2024survey}, but low-precision training is not their central scope. System-oriented surveys emphasize parallelism, scheduling, memory management, and communication infrastructure, whereas low-bit LLM surveys often place training beside inference-time compression and deployment. In contrast, training-time low precision must also account for gradients, optimizer states, activation storage, and stability mechanisms that have no direct inference-only counterpart. This gap motivates a survey that treats low precision as both a representation and optimization-pipeline problem.

\begin{figure*}[t]
    \centering
    \includegraphics[width=0.95\linewidth]{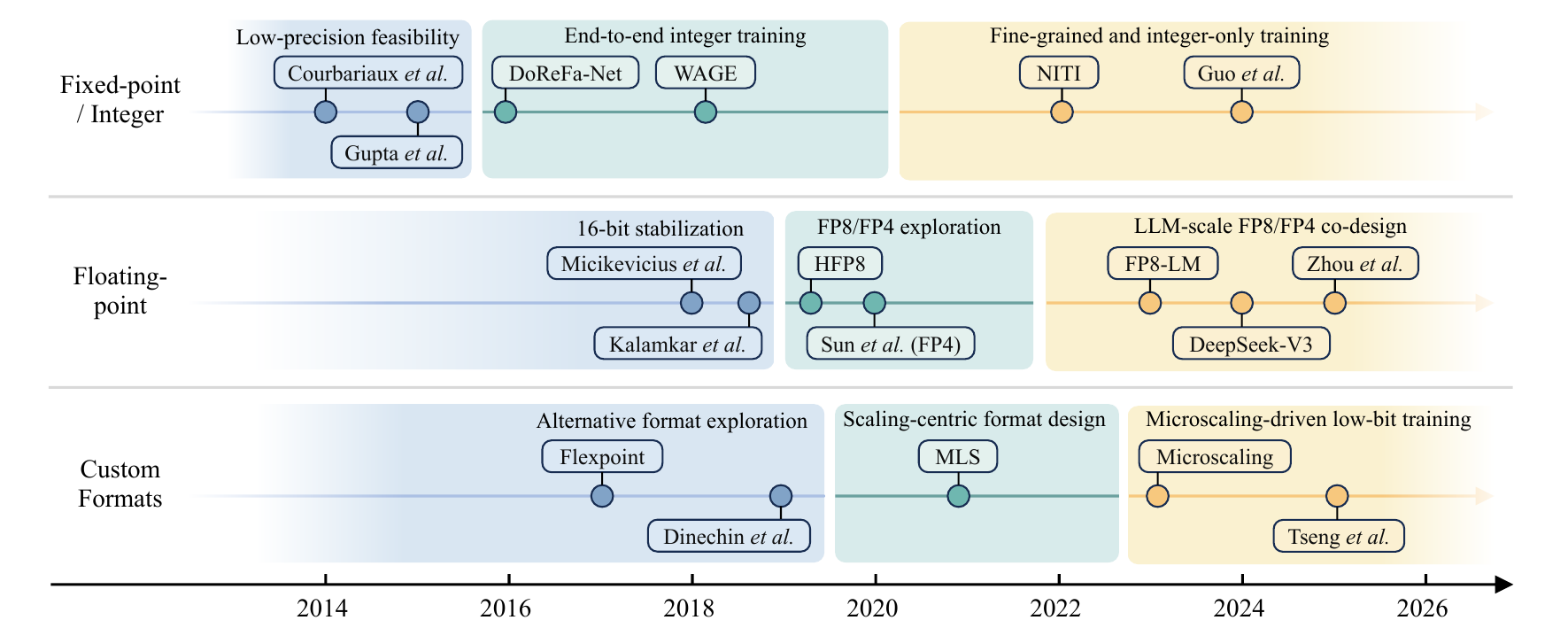}
    \caption{Evolution of low-precision training across the major numerical-format families, highlighting development stages from selected works. The figure is intended as an illustrative summary and does not provide a complete chronology of all studies.}
    \label{fig:timeline}
\end{figure*}

In this survey, we focus on low-precision training of models and provide a comprehensive, structured review of recent advances in the field. We organize the literature primarily by value representation formats, which shape hardware design, computational performance, and the way readers navigate the field. To avoid a purely format-based list, we also use four recurring challenges as a reading guide: (1) dynamic-range mismatch and outlier handling, (2) accumulation and update-fidelity protection, (3) precision allocation across tensors, layers, operators, and training stages, and (4) system/dataflow co-design for efficient execution and communication. These challenges help explain why methods with different encodings often converge to similar remedies, such as finer-grained scaling, selective high-precision paths, protected optimizer states, or cast-aware kernels. This cross-cutting view supports family-level efficiency--accuracy comparison, controlled benchmarking, and discussion of robustness and model-reliability implications. In addition, we also cover Quantization-Aware Training (QAT) techniques and system-level solutions that enable efficient low-precision training.
Specifically, the topics covered in this survey are organized as follows:
\begin{itemize}
    \item \textbf{Fixed-point and integer-based methods.} Fixed-point and integer representations are the most widely adopted numerical formats in low-precision training. Early works primarily relied on fixed-point quantization with global scaling factors. More recent studies have shifted toward integer quantization with finer-grained scalers to better handle outliers and improve precision.
    \item \textbf{Floating-point-based methods.} Floating-point formats represent another mainstream approach in low-precision training. While they share a similar quantization process with fixed-point methods, recent advances in hardware support have led to increased interest in leveraging low-precision floating-point numbers, enabling both flexibility and improved performance.
    \item \textbf{Customized format-based methods.} Beyond standard numerical formats, various customized representations have been proposed to further optimize performance. These formats are often derived from standard types.
    \item \textbf{Quantization-aware training methods.} Unlike general low-precision training, which applies quantization to weights, activations, and gradients, QAT focuses primarily on simulating quantization effects during the forward pass. Some methods rely on fake quantization to mimic quantization noise without improving training efficiency, while others incorporate real quantization, offering partial gains in training efficiency.
    \item \textbf{System support.} The practical adoption of low-precision training heavily depends on system-level support. This section discusses infrastructures that enable or accelerate low-precision training workflows.
\end{itemize}
\begin{figure}[t]
    \centering
    \includegraphics[width=\linewidth]{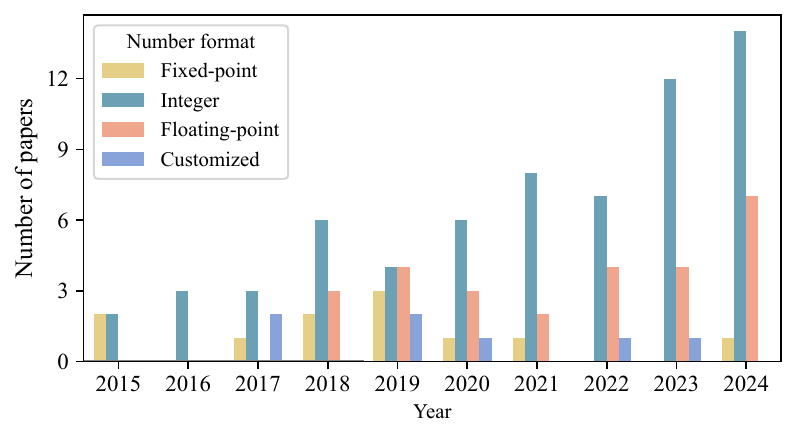}
    \caption{Annual count of reviewed papers in this survey from 2015 to 2024, categorized by primary adopted numerical format. The statistics are limited to papers published up to 2024.}
    \label{fig:trend}
\end{figure}
We also identify key open challenges and highlight promising directions for future research. Overall, this survey aims to offer a clear and comprehensive understanding of the field, laying the groundwork for more efficient and scalable training practices in the era of large language models.
Figure~\ref{fig:trend} compares the papers reviewed in this survey with different low-precision representations up to 2024, revealing a clear upward trend in integer-based methods over the past decade, alongside a recent rise in floating-point methods. In contrast, fixed-point representations have experienced a steady decline, while customized formats have consistently garnered minimal interest.
Moreover, Figure~\ref{fig:timeline} summarizes how low-precision training has evolved across the three main numerical-format families discussed in this survey. It highlights representative milestones and stage-wise shifts in technical emphasis.

The main contributions of this survey are as follows:
\begin{itemize}
    \item We provide a detailed classification of low-precision training techniques, organized by numerical formats, including fixed-point, integer, floating-point, and customized representations. Different from prior surveys that treat low precision as part of broader efficient-training, low-bit LLM, or deployment-oriented compression landscapes, this work takes training-time low-precision methods as the central scope.
    \item Beyond categorization, we synthesize recurring challenges across format families, including range management, update fidelity, precision allocation, and system/dataflow co-design. We also summarize representative methods in tabular form, compare family-level efficiency--accuracy trade-offs with qualitative synthesis and controlled benchmarking, and discuss robustness and model-reliability implications.
    \item We examine QAT techniques that closely resemble low-precision training and provide an overview of typical works that apply QAT in the context of LLM training. Moreover, system supports that enable efficient low-precision training are also introduced.
    \item We outline key challenges in the field and discuss promising directions for future research aimed at developing more efficient and environmentally sustainable training practices for large models.
\end{itemize}

The survey is organized as follows. As shown in Figure~\ref{fig:structure}, Section~\ref{sec:bg} provides background on low-precision training, including value representation formats and affected components during training. Section~\ref{sec:fx} discusses techniques based on fixed-point and integer formats, while Section~\ref{sec:fp} covers floating-point approaches. Section~\ref{sec:custom} presents customized low-precision formats. Section~\ref{sec:cross} then revisits the literature from cross-cutting perspectives, making a family-level comparison of memory, speed, and accuracy as well as robustness and model-reliability implications. Section~\ref{sec:qat} discusses QAT methods, and Section~\ref{sec:system} introduces system-level support strategies. Section~\ref{sec:future} outlines open problems and research opportunities, and Section~\ref{sec:conclusion} concludes the survey.

\input{fig/forest_structure}

\input{contents/sec_bg}

\input{contents/sec_fx}

\input{contents/sec_fp}

\input{contents/sec_custom}

\input{contents/sec_cross}

\input{contents/sec_qat}

\input{contents/sec_system}

\input{contents/sec_future}

\input{contents/sec_conclusion}

\bibliographystyle{IEEEtran}
\bibliography{ref}

\clearpage

\appendices

\input{contents/app_fx_details}

\input{contents/app_fp_details}

\input{contents/app_custom_details}

\input{contents/app_qat_details}

\end{document}

%% file: fig/forest_structure.tex
\usetikzlibrary{fit}

\tikzstyle{leaf}=[
    align=center,
    inner xsep=10pt,
    inner ysep=3pt,
]

\begin{figure}[htbp]
    \centering
    \begin{forest}
        forked edges,
        for tree={
            font=\sffamily\bfseries, 
            text=textcolor, 
            grow=east,
            reversed=true,
            anchor=base west,
            parent anchor=east,
            child anchor=west,
            base=center,
            rectangle,
            draw=black,
            rounded corners,
            align=left,
            text centered,
            minimum width=4em,
            edge+={black, line width=1pt},
            s sep=3pt,
            inner xsep=2pt,
            inner ysep=3pt,
            line width=0.8pt,
            ver/.style={rotate=90, child anchor=north, parent anchor=south, anchor=center},
            myline/.style={line width=0.8pt},
        },
        where level=0{fill=rootcolor,text=white,text width=9em,font=\footnotesize,}{},
        where level=1{fill=level3color,text width=18em,font=\footnotesize,leaf,}{},
        [
        {Low-Precision Training}, ver,
                [
                {Basic Knowledge (\S~\ref{sec:bg})}, name=bg
                ]
                [, phantom, inner ysep=2pt]
                [
                {Low-Precision Fixed-Point and Integer Training (\S~\ref{sec:fx})}, name=fx
                ]
                [
                {Low-Precision Floating-Point Training (\S~\ref{sec:fp})}, name=fp
                ]
                [
                {Custom Numerical Formats (\S~\ref{sec:custom})}, name=custom
                ]
                [
                {Cross-Cutting Perspectives (\S~\ref{sec:cross})}, name=cross
                ]
                [
                {Quantization-Aware Training Techniques (\S~\ref{sec:qat})}, name=qat
                ]
                [
                {System Support (\S~\ref{sec:system})}, name=system
                ]
                [, phantom, inner ysep=6pt]
                [
                {Future Directions (\S~\ref{sec:future})}, name=future
                ]
                [
                {Conclusion (\S~\ref{sec:conclusion})}, name=conclusion
                ]
        ]
        \node[draw=black, dashed, rounded corners, inner sep=3pt, line width=0.8pt, fit=(fx) (fp) (custom) (cross) (qat) (system), 
        label={[font=\footnotesize, text=textcolor]below: Core Survey Body}] (box) {};
    \end{forest}
    \caption{Overview of the survey structure and key components.}
    \label{fig:structure}
\end{figure}

%% file: contents/sec_bg.tex
\section{Basic Knowledge}
\label{sec:bg}

\begin{figure}[htpb]
  \centering
  \includegraphics[width=\linewidth]{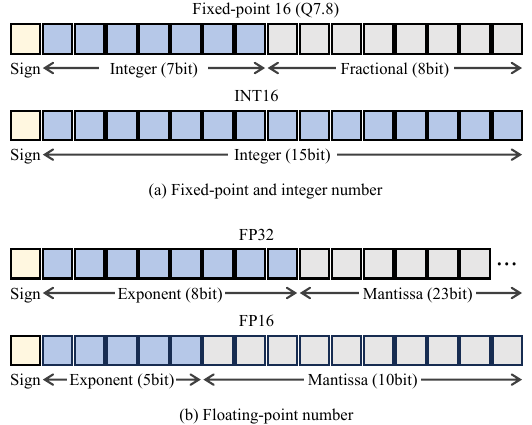}
  \caption{Visualization of different value representation formats.}
  \label{fig:number}
\end{figure}

This section introduces the value representation formats used throughout the survey and the training components commonly targeted by low-precision methods. Figure~\ref{fig:number} visualizes representative fixed-point, integer, and floating-point formats.

\subsection{Value Representation Formats}

\textbf{Fixed-Point and Integer Formats.}
Fixed-point representation assigns fixed bit counts to the integer and fractional parts of a value. Without fractional bits, it reduces to conventional integer representation. A signed fixed-point scalar $a$ with bit-width $B$ and binary representation $R = (a_0, a_1, \dots, a_{B-1}) \in \{0,1\}^{B}$ can be written as~\mcite{Sakr}{sakrs19}:
\begin{equation}
  a = r \left( -a_0 + \sum_{i=1}^{B-1} 2^{-i} a_i \right),
  \label{eq:fx}
\end{equation}
where $r$ denotes the predefined dynamic range and is often restricted to a power of two for hardware simplicity. When $r = 2^{B-1}$, the expression becomes the standard two's-complement integer representation:
\begin{equation}
  a = -2^{B-1} a_0 + \sum_{i=1}^{B-1} 2^{B-1-i} a_i.
\end{equation}
In this case, the fractional component vanishes and the result lies in $[-2^{B-1}, 2^{B-1}-1]$. Fixed-point arithmetic is hardware-efficient, but its static precision requires careful control of overflow and underflow.

\textbf{Floating-Point Formats.}
Floating-point representations are widely used because they support a much larger dynamic range. A floating-point number contains a sign bit $s$, an exponent $e$ that controls range, and a mantissa $m$ that controls precision. Its general form is:
\begin{equation}
    x = (-1)^s \times m \times 2^{e - b},
    \label{eq:fl}
\end{equation}
where $b$ denotes the exponent bias.

The IEEE 754 standard~\cite{ieee754} defines widely used formats such as FP32 single precision (E8M23)\footnote{FP32 denotes a 32-bit floating-point number, with 8 bits allocated for the exponent $e$ and 23 bits for the mantissa $m$.} and FP16 half precision (E5M10). Deep learning has also introduced specialized formats such as BF16 (E8M7) and FP8 (E4M3, E5M2), which offer favorable trade-offs between precision, range, and computational efficiency and now underpin many mixed-precision training recipes~\cite{micikeviciusnad18,kalamkar2019study,micikevicius2022fp8}.
In pursuit of even more compact representations, recent hardware and software stacks are also exploring ultra-low-precision formats such as FP4 and microscaling-based datatypes to further reduce memory traffic and arithmetic cost~\cite{rouhani2023microscaling,tseng2025training}.

\textbf{Customized Formats.}
Beyond standard fixed-point and floating-point representations, there also exist customized numerical formats specifically designed for deep learning applications, such as Microscaling~\cite{rouhani2023microscaling}. These specialized formats will be introduced in context as we discuss the corresponding works in Section~\ref{sec:custom}.

\subsection{Different Low-Precision Components}

\begin{figure*}[htpb]
  \centering
  \includegraphics[width=\linewidth]{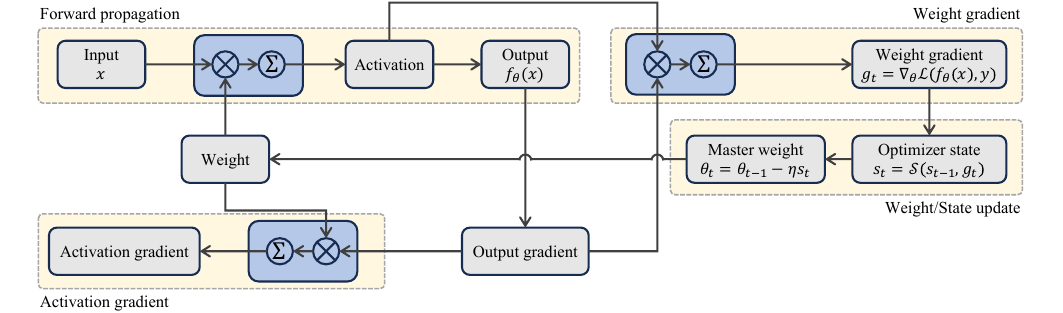}
  \caption{Generalized model training pipeline, showcasing components that can be optimized with low-precision methods. }
  \label{fig:pipe}
\end{figure*}

Several training components can be optimized with low-precision arithmetic. We summarize the training dynamics as:
\begin{equation}
    \begin{aligned}
    \theta_t &= \theta_{t-1} - \eta \cdot s_t, \\
    s_t &= \mathcal{S}(s_{t-1}, g_t), \quad &\text{(State update)} \\
    g_t &= \nabla_\theta \mathcal{L}(f_\theta(x), y), \quad &\text{(Gradient computation)}
    \end{aligned}
\end{equation}
where $\theta_t$ denotes the model parameters at iteration $t$, $\eta$ is the learning rate, and $s_t$ denotes optimizer state such as momentum or adaptive estimates in Adam~\cite{adam}. The function $\mathcal{S}$ defines the optimizer update, and $g_t$ is the loss gradient $\mathcal{L}$ over a mini-batch $(x, y)$. Figure~\ref{fig:pipe} visualizes this abstraction.

Prior work has primarily focused on applying low-precision optimization to one or more components of the training process. A straightforward approach involves quantizing the model weights $\theta_t$, activations used in the forward pass $f_\theta(x)$, and gradients $g_t$ computed during backpropagation. In addition, some studies aim to further reduce the memory footprint by representing stored activations and optimizer state $s_t$ in low precision~\cite{micikeviciusnad18,chakrabartim19,dettmers20228bit,peng2023fp8}.

%% file: contents/sec_fx.tex
\section{Low-Precision Fixed-Point and Integer Training}
\label{sec:fx}

In this section, we review low-precision training methods based on fixed-point and integer quantization. Earlier work often used these terms interchangeably. Here we use fixed-point when the scaling factor $\Delta$ is constant and integer quantization when it is data-dependent. For a $B$-bit representation with $k$ fractional bits ($k < B$), the quantized value $q$ of input $x$ is:
\begin{equation}
\begin{aligned}
q &= \text{clip}\left( \text{round}\left( \frac{x}{\Delta} \right), -2^{B-1}, 2^{B-1}-1 \right), \\
\Delta &= 2^{-k}.
\end{aligned}
\end{equation}
This representation is hardware-efficient but limited in dynamic range. As accelerator support for integer arithmetic improved, research shifted toward integer quantization. A common choice is symmetric uniform quantization~\cite{krishnamoorthi2018quantizing}, where the scaling factor $\Delta$ is data-dependent. Given an input range $(l, u)$, the quantized value is computed as:
\begin{equation}
\begin{aligned}
q &= \text{round}\left( \frac{\text{clip}(x, -c, c)}{\Delta} \right), \\
\Delta &= \frac{c}{2^{B-1} - 1}, \\
c &= \max(|l|, |u|).
\end{aligned}
\end{equation}
In both quantization formats, the original input is approximated by reconstructing the value $\hat{x}$ as:
\begin{equation}
\hat{x} = q \cdot \Delta.
\end{equation}

Apart from the reduced precision of values, training largely follows the full-precision forward and backward passes. The main exception is 1-bit training, which commonly uses the Straight-Through Estimator (STE) to approximate gradients through the non-differentiable quantizer.

Rather than following the chronology of individual papers, we organize this section around the design axes that shaped fixed-point and integer training. 
Fixed-point methods are discussed through the trade-off between arithmetic simplicity, range limitation, rounding, and adaptive scaling. Integer methods are grouped by end-to-end integerization, range and outlier control, update fidelity, sensitive components, optimizer-state compression, and communication limits.

\subsection{Fixed-Point Training}

Fixed-point training is best understood as a trade-off between simple arithmetic and limited dynamic range. Its storage and arithmetic are attractive for hardware, but a static or coarse scale must cover tensors whose magnitudes vary across layers, operators, and training stages.
The representative fixed-point methods covered in this section are summarized in Table~\ref{tab:fixed_methods}.

\input{contents/tables/tab_fixed_methods}

\textbf{Range and rounding.}
Coarse fixed-point quantization can erase small optimization signals. When the quantization step is too large, weak gradients or parameter updates may be rounded to zero, making the optimizer insensitive to accumulated learning signals. Stochastic rounding addresses this problem by making the rounded value unbiased in expectation:
\begin{align}
	R(x)= \begin{cases}
		\lfloor x \rfloor   &\text{with probability}~p(x),\\
		\lfloor x \rfloor+\delta   &\text{with probability}~1-p(x),
	\end{cases}
\end{align}
with $p(x) = 1 - \frac{x - \lfloor x \rfloor}{\delta}$. Because this rule satisfies $\mathcal{E}(R(x)) = x$, it can preserve weak gradient signals that nearest rounding would remove~\cite{guptaagn15}. Later variants simplify the stochastic rule to reduce implementation cost while retaining the same goal of protecting sub-step updates~\cite{xia2021simple}. This line of work shows that fixed-point training is not determined only by bit-width; the rounding rule also controls whether small but consequential updates survive.

\textbf{Adaptive scaling.}
Range adaptation becomes necessary because a single fixed scale cannot simultaneously represent large activations, small gradients, and layer-wise distribution shifts. Fixed-point methods therefore evolved from ordinary shared-scale formats toward adaptive layer-wise scaling, range tracking, trainable binary-point placement, and precision schedules~\cite{chenhzx17,zhanglzlhzggdzc20,rajagopalvvb20,dai2024trainable}. Related designs reduce memory pressure through low-precision SGD, compressed activation storage, or averaging-based optimization while limiting the instability caused by coarse quantization~\cite{sakrs19,chakrabartim19,yangzkbws19}. Dynamic fixed-point methods further relax the static-scale assumption by using multiple adjustable scaling factors, often combined with clipping, batch-size scaling, or architecture-specific recipes for larger workloads~\cite{williamson1991dynamically,courbariaux2014training,joplc18,0002mmkab0vkghd18}.

These techniques differ in implementation, but they share the same design logic: keep the arithmetic close to fixed-point while allowing the scale to follow the tensor distribution. The benefit is that the hardware path remains simpler than full floating-point arithmetic. The cost is that scale selection becomes another control problem, and incorrect updates can still produce overflow, underflow, or excessive rounding error.

Overall, fixed-point studies established the feasibility of low-precision training under simple arithmetic, but they also revealed the cost of shared-scale rigidity. Fixed-point formats remain attractive when hardware simplicity and low storage overhead dominate, yet are fragile under outliers, layer-wise distribution shifts, and heterogeneous tensor statistics. This helps explain why later work increasingly moved toward finer-grained integer, floating-point, or custom-format designs.

\subsection{Integer Training}

Integer training generalizes the fixed-point motivation but shifts the central question from whether a simple scale can work to which tensors, operators, states, and communication paths require distinct scaling or protection.
This framing keeps the discussion format-specific while making the analytical progression explicit.
Representative methods are summarized in Table~\ref{tab:integer_methods}.

\input{contents/tables/tab_integer_methods}

\textbf{End-to-end integerization.}
End-to-end integerization asks whether the main training path can use integer values rather than restricting quantization to inference. Early integer-training frameworks demonstrated that weights, activations, and parameter gradients can all be quantized during training, opening the way to bit-level acceleration~\cite{zhou2016dorefa,wulcs18}. Fully quantized extensions broadened this idea to more of the pipeline, including optimizer-related quantities~\cite{yangdwyxl20}. More recent studies show that low-bit integer training can remain viable, but usually only when outliers, master-weight storage, or fragile tensors are handled explicitly~\cite{zhang2025accurate,zhao2024direct}.

Together, these studies shift the question from whether integer training is possible to where scaling, compensation, or high-precision exceptions are needed. Thus, integer methods should be compared not only by how many tensors they quantize, but also by which parts of the pipeline remain numerically protected for stable, efficient training.

\textbf{Range and granularity.}
Range and outlier control addresses tensors with highly non-uniform distributions. A few large activations or gradients can determine the scale for many smaller values, causing either outlier clipping or loss of resolution. One family of methods redesigns the quantizer through logarithmic quantization, variance-aware or history-based range adaptation, and gradient-specific schemes aligned with channel-wise or projected structure~\cite{miyashita2016convolutional,de2018high,zhaohplzgx21,fournarakisn21,zhang2024q}. Another family transforms the data before quantization: Hadamard-domain and related structured representations suppress activation outliers and reduce backward-pass cost in low-bit transformer or continual-learning settings~\cite{xilcz23,schiemer2023hadamard,kim2024hlq}.

As integer training matured, precision also became a resource to allocate rather than a uniform setting. Layer-aware precision adaptation, per-layer dynamic-range control, dynamic bit allocation, and block-wise or capacity-aware schemes match scales and bit-widths to heterogeneous tensor sensitivity~\cite{zhanglzlhzggdzc20,wangrls22,dingfhddlzx24,shenlslgll25,fuyz0lgwl20,xicztcz24}. The common trade-off is clear: global scaling is simple but fragile, whereas fine-grained scaling improves accuracy at the cost of metadata, casting, and kernel-design overhead.

\begin{figure}[t]
    \centering
    \includegraphics[width=\linewidth]{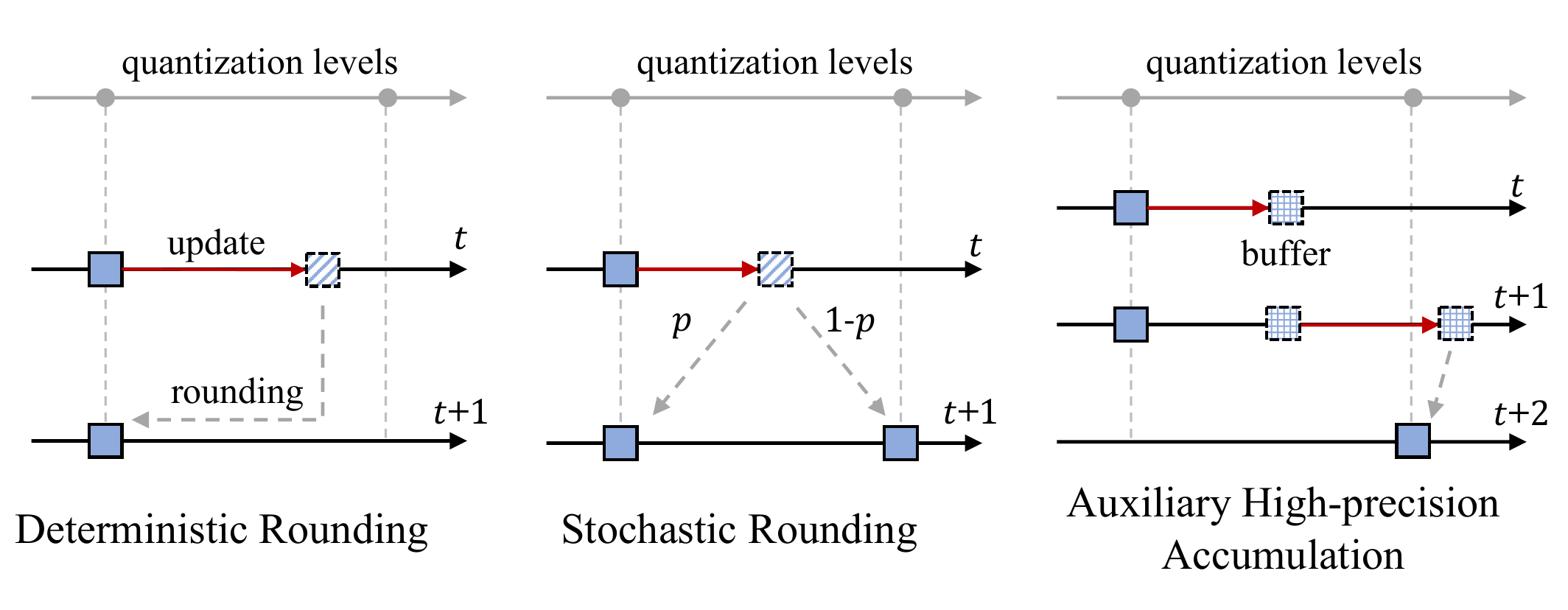}
    \caption{Illustration of how sub-step updates can be lost in low-precision training. Deterministic rounding maps a small update back to the original quantization level, stochastic rounding preserves it in expectation by probabilistically selecting adjacent levels, and auxiliary high-precision accumulation stores small updates until they trigger a discrete low-precision change.}
    \label{fig:update}
\end{figure}

\textbf{Update fidelity.}
Update fidelity determines whether the optimization trajectory is preserved after discretization. Small updates can disappear once they fall below the quantization step, and repeated rounding can distort gradients even when forward activations remain accurate. Some methods preserve selected operations in elevated precision, whereas others use auxiliary accumulation buffers to recover updates that would otherwise be lost~\cite{bannerhhs18,park2018training}.
These studies point to a common failure mode: once updates become smaller than the quantization step, naive low-precision arithmetic can silently erase them. Figure~\ref{fig:update} abstracts this recurring issue. When a true update falls between adjacent quantization levels, deterministic rounding maps it back to the original low-precision value, whereas stochastic rounding or an auxiliary high-precision buffer preserves the update either in expectation or through delayed accumulation.

Later methods reduce gradient quantization error through improved grouping, clipping, and error modeling~\cite{zhugylwlyy20,guo2024towards}. A related strategy is to vary precision over training time: early stages can often tolerate coarser precision, whereas later stages require more accurate updates near convergence~\cite{fuyz0lgwl20,fuglydcl21}.

Sensitive operators reinforce the same lesson. Normalization layers are especially fragile because their statistics, divisions, and rescaling operations interact poorly with coarse integer arithmetic. Existing methods either preserve such operators in higher precision or redesign them with range-based, constant-scaling, L1-based, or quantized L1 formulations~\cite{zhou2016dorefa,yangcdygl22,bannerhhs18,yangdwyxl20,guo2024towards,yangdyxl21}. Integer recipes have also been extended beyond standard image classification to segmentation and vision-language model training, where task-specific modules again require component-aware precision choices~\cite{yangdyxl21,wortsmandzmfs23}. These examples show that the same bit-width can affect convolutional layers, transformer projections, normalization, optimizer updates, and task-specific heads differently.

Theoretical work complements these empirical recipes by explaining when integer training can remain reliable. 
Existing analyses show that training is often more quantization-robust than naive worst-case reasoning suggests, while distinguishing convex from non-convex settings, dimension-dependent from dimension-free bounds, and naive from statistically designed fully quantized training~\cite{bannerhhs18,lid0ssg17,lis19,zhugylwlyy20,zhang0kalz17,chengymg20}.

\textbf{Optimizer and communication compression.}
Optimizer states and communication paths are auxiliary to arithmetic kernels but can dominate training cost. For adaptive optimizers such as Adam~\cite{adam}, first- and second-order statistics can require 2-3$\times$ the memory of SGD. Block-wise schemes and range adaptation can reduce Adam moments to 16-bit or 8-bit, but more aggressive compression must handle rare outliers, zero-point effects, and error accumulation~\cite{rameshpggvrcs21,dettmerslsz22,chitsazfmc24,modoranusmk0ra24,licz23}. Beyond Adam, integer compression has been explored for optimizers such as Lion and Shampoo through structure-aware state quantization and low-bit representations for second-order preconditioners or their decomposed forms~\cite{li2023qft,li2024memory,wangzlh24}. Optimizer compression is therefore constrained less by average precision than by rare but consequential state outliers.

Activation storage and dataflow overhead form another bottleneck. Activation-compression methods reduce backward-pass memory by exploiting narrow activation ranges with sparse outliers, adaptive per-group compression, runtime control, and nonlinear backward approximations~\cite{parkyv18,chenzywsm021,liuzwcchcltgmc22,novikovbgsdo23,liuzylch22,eliassens24}. For transformer training, repeated quantize-compute-dequantize conversion can also weaken practical speedups. Systems such as Jetfire redesign operators to support more direct INT8 data flow, while methods addressing latent full-precision weight paths reduce memory that would otherwise remain outside the quantized arithmetic path~\cite{xicztcz24,feidzzlzx25}.

At large scale, communication rather than arithmetic can dominate training cost. Communication-oriented integer methods compress values exchanged across devices while controlling compression bias. Early frameworks such as QSGD established convergence guarantees for quantized communication~\cite{dettmers20158,alistarhg0tv17}. Later methods reduce bias through error feedback, bidirectional compensation, immediate local error compensation, adaptive quantizers, joint quantization-and-sparsification, and preprocessing that avoids decompression overhead~\cite{tangylzl19,chenshl21,chensll23,cheng2025communication,xie2025loco,yuwh19,faghritma0r20,libvlxmy24}. Modern large-model systems further incorporate block quantization into optimizer sharding, pipeline parallelism, ShardedDP, and FSDP~\cite{wang2023zero++,wangyrhdcr022,jiaxlwf0slzlt24,markovvga23}. Here, the main contribution is often a better match between quantization and the actual memory or communication bottleneck rather than a new integer format alone.

Overall, integer training has progressed from uniform low-bit quantization toward selective schemes that protect outliers, sensitive layers, optimizer states, and expensive data movement. Practical fully quantized training rarely comes from quantizing everything identically; it comes from matching scale granularity, update protection, high-precision exceptions, and system dataflow to tensor statistics and operator roles.

\textbf{Binary and 1-Bit training.}
Binary training represents the extreme limit of the integer family. Binary neural networks (BNNs) quantize weights and activations to $\pm1$, replacing multiply-accumulate operations with simpler arithmetic. The core difficulty is training discrete weights while still accumulating meaningful updates, so a common solution keeps full-precision latent weights for gradient accumulation while using binary values during propagation~\cite{courbariauxbd15}. Its binarization can be implemented either deterministically with the sign function,
\begin{align}
	q = \mathrm{sign}(x) =  \begin{cases}
		+1   &\text{if}~x \geq 0,\\
		-1   &\text{otherwise},
	\end{cases}
\end{align}
or stochastically as
\begin{align}
	q = \begin{cases}
		+1   &\text{with probability}~\sigma(x),\\
		-1   &\text{with probability}~1-\sigma(x),
	\end{cases}
\end{align}
where $\sigma(x)$ denotes the hard sigmoid function~\cite{courbariauxbd15}. Related binary architectures further binarize layer inputs and introduce scaling-factor approximations, showing that aggressive binarization can deliver large memory and convolution-cost savings~\cite{rastegariorf16}. Subsequent work quantizes gradients, stores activations directly in binary form, and analyzes how gradient variance affects fully quantized 1-bit training~\cite{zhou2016dorefa,wangdmzlcccc23,gao20241}.

A parallel line uses 1-bit quantization for communication-efficient distributed optimization rather than full binary network training. Error feedback preserves residual information lost at each compression step and can make compressed training approach full-precision SGD behavior~\cite{seidefdly14,karimireddyrsj19}. Later methods extend this principle to adaptive optimizers and large-batch training through staged strategies, frozen or partially frozen preconditioners, compressed momentum updates, and sign-based communication frameworks for practical distributed settings~\cite{tanggarlllzh21,lulzsh23,liatrh22,wuh0qwz022,pengqywwl23}. Recent theory also analyzes sign-based nonconvex acceleration under weaker smoothness assumptions and compressed distributed protocols~\cite{sun2023rethinking}.

It is therefore useful to distinguish binary network training from 1-bit communication. The former changes the representational capacity of the model, whereas the latter primarily compresses exchanged optimization signals. Both mark the efficiency frontier, but they require hidden high-precision paths, residual compensation, staged optimization, or selective exceptions to protect optimization-critical information.

%% file: contents/tables/tab_fixed_methods.tex
\begin{table*}[t]
	\centering
	\caption{Representative fixed-point training methods.}
	\setlength{\tabcolsep}{8pt}
	\begin{tabular}{lllllcccccc}
		\toprule
		Paper                                        & Family & Model   & Metrics & Weights   & Activations & Gradients    & Optim.  & Comm. & Grouping \\
		\midrule
		\mcite{Gupta}{guptaagn15}                    & FX     & MLP/CNN & Acc     & fixed16   & fixed16     & fixed16      & -       & -     & -        \\
		\mcite{Xia}{xia2021simple}                   & FX     & MLP     & Acc     & fixed16   & -           & fixed16      & -       & -     & -        \\
		FxpNet~\cite{chenhzx17}                      & FX     & CNN     & Acc     & fixed12   & binary      & fixed12      & fixed12 & -     & -        \\
		\mcite{Courbariaux}{courbariaux2014training} & FX     & Maxout  & Acc     & fixed10   & fixed10     & fixed12      & -       & -     & -        \\
		\mcite{Zhang}{zhanglzlhzggdzc20}             & FX     & CNN/RNN & Acc/PPL & fixed8    & fixed8      & fixed8/16/24 & -       & -     & -        \\
		MuPPET~\cite{rajagopalvvb20}                 & FX     & CNN     & Acc     & fixed16   & fixed16     & fixed16      & -       & -     & -        \\
		QFX~\cite{dai2024trainable}                  & FX     & CNN/BNN & Acc     & learnable & learnable   & learnable    & -       & -     & -        \\
		\bottomrule
	\end{tabular}
	\label{tab:fixed_methods}
\end{table*}

%% file: contents/tables/tab_integer_methods.tex
\begin{table*}[t]
	\centering
	\caption{Representative integer and multi-bit training methods, including binary and 1-bit variants.}
	\setlength{\tabcolsep}{5pt}
	\begin{tabular}{lllllcccccc}
		\toprule
		Paper                               & Family       & Model           & Metrics & Weights   & Activations & Gradients & Optim. & Comm. & Grouping \\
		\midrule
		\mcite{Das}{0002mmkab0vkghd18}      & Integer      & CNN             & Acc     & INT16     & INT16       & INT16     & -      & -     & -        \\
		WAGE~\cite{wulcs18}                 & Integer      & DNN             & Acc     & INT2      & INT8        & INT8      & -      & -     & -        \\
		WAGEUBN~\cite{yangdwyxl20}          & Integer      & CNN             & Acc     & INT8      & INT8        & INT8      & INT8   & -     & -        \\
		NITI~\cite{wangrls22}               & Integer      & DNN             & Acc     & INT8      & INT8        & INT8      & -      & -     & -        \\
		SWALP~\cite{yangzkbws19}            & Integer      & CNN             & Acc     & INT8      & INT8        & INT8      & INT8   & -     & -        \\
		\mcite{Banner}{bannerhhs18}         & Integer      & CNN             & Acc     & INT8      & INT8        & INT8      & -      & -     & -        \\
		\mcite{Zhao}{zhaohplzgx21}          & Integer      & CNN             & Acc     & INT8      & INT8        & INT8      & -      & -     & -        \\
		In-hindsight~\cite{fournarakisn21}  & Integer      & CNN             & Acc     & INT8      & INT8        & INT8      & -      & -     & -        \\
		\mcite{Zhu}{zhugylwlyy20}           & Integer      & CNN             & Acc     & INT8      & INT8        & INT8      & -      & -     & -        \\
		AMPA~\cite{dingfhddlzx24}           & Integer      & CNN             & Acc     & INT4-8    & INT4-8      & INT4-8    & -      & -     & -        \\
		\mcite{Shen}{shenlslgll25}          & Integer      & CNN             & Acc     & INT8      & INT8        & INT8      & -      & -     & -        \\
		FracTrain~\cite{fuyz0lgwl20}        & Integer      & CNN             & Acc     & INT4-8    & INT4-8      & INT6-12   & -      & -     & -        \\
		CPT~\cite{fuglydcl21}               & Integer      & CNN/LSTM        & Acc/PPL & INT3-8    & INT3-8      & -         & -      & -     & -        \\
		\mcite{Chen}{chengymg20}            & Integer      & CNN             & Acc     & INT5      & INT5        & INT5      & -      & -     & block    \\
		\mcite{Park}{parkyv18}              & Act. Comp.   & CNN             & Acc     & -         & INT3        & -         & -      & -     & -        \\
		\mcite{Xi}{xilcz23}                 & Integer      & Transformer     & PPL/Acc & INT4      & INT4        & INT4      & -      & -     & -        \\
		\mcite{Zhang}{zhang2025accurate}    & Integer      & LLM             & PPL/Acc & INT8      & INT8        & INT8      & -      & -     & block    \\
		Jetfire~\cite{xicztcz24}            & Integer      & LLM             & PPL/Acc & INT8      & INT8        & INT8      & -      & -     & block    \\
		\mcite{Chitsaz}{chitsazfmc24}       & Integer      & LLM             & PPL/Acc & INT4/INT8 & INT4/8      & INT4/8    & INT4/8 & -     & -        \\
		SwitchBack~\cite{wortsmandzmfs23}   & Integer      & VLM             & Acc     & INT8      & INT8        & INT8      & -      & -     & -        \\
		\mcite{Guo}{guo2024towards}         & Integer      & CNN/Transformer & Acc/PPL & INT4/INT6 & INT4/6      & INT4/6    & -      & -     & -        \\
		HDQT~\cite{schiemer2023hadamard}    & Integer      & CNN             & Acc     & INT4      & INT4        & INT4      & -      & -     & -        \\
		\mcite{Yang}{yangdyxl21}            & Integer      & CNN             & mIoU    & INT8      & INT8        & INT8      & -      & -     & -        \\
		DQT~\cite{zhao2024direct}           & Integer      & LLM             & PPL/Acc & INT8      & -           & -         & -      & -     & -        \\
		Q-GaLore~\cite{zhang2024q}          & Optim. Comp. & LLM             & PPL/Acc & INT8      & -           & INT4      & -      & -     & layer    \\
		QFT~\cite{li2023qft}                & Optim. Comp. & LLM             & Acc     & -         & -           & -         & INT8   & -     & channel  \\
		\mcite{Dettmers}{dettmerslsz22}     & Optim. Comp. & CNN/Transformer & Acc/PPL & -         & -           & -         & INT8   & -     & block    \\
		\mcite{Wang}{wangzlh24}             & Optim. Comp. & CNN/Transformer & Acc/PPL & -         & -           & -         & INT4   & -     & -        \\
		\mcite{Li}{li2024memory}            & Optim. Comp. & CNN/ViT         & Acc     & -         & -           & -         & INT4   & -     & -        \\
		\mcite{Li}{licz23}                  & Optim. Comp. & CNN/Transformer & Acc/PPL & -         & -           & -         & INT4   & -     & row/col. \\
		\mcite{Chakrabarti}{chakrabartim19} & Act. Comp.   & CNN             & Acc     & -         & INT4        & -         & -      & -     & -        \\
		ActNN~\cite{chenzywsm021}           & Act. Comp.   & CNN             & Acc     & -         & INT2        & -         & -      & -     & -        \\
		GACT~\cite{liuzwcchcltgmc22}        & Act. Comp.   & CNN/Transformer & Acc/PPL & -         & INT1-7      & -         & -      & -     & tensor   \\
		\mcite{Novikov}{novikovbgsdo23}     & Act. Comp.   & CNN             & Acc     & -         & -           & INT1-3    & -      & -     & -        \\
		EXACT~\cite{liuzylch22}             & Act. Comp.   & GNN             & Acc/F1  & -         & INT2        & -         & -      & -     & -        \\
		\mcite{Eliassen}{eliassens24}       & Act. Comp.   & GNN             & Acc/F1  & -         & INT2        & -         & -      & -     & block    \\
		HLQ~\cite{kim2024hlq}               & Act. Comp.   & CNN/Transformer & Acc/PPL & -         & -           & INT4      & -      & -     & -        \\
		\mcite{Dettmers}{dettmers20158}     & Comm. Comp.  & CNN             & Acc     & -         & -           & -         & -      & W8A8  & -        \\
		QSGD~\cite{alistarhg0tv17}          & Comm. Comp.  & DNN/CNN         & Acc/WER & -         & -           & -         & -      & G4/G8 & -        \\
		LIEC-SGD~\cite{cheng2025communication} & Comm. Comp.  & DNN             & Acc     & -         & -           & -         & -      & G comp. & -     \\
		ZeRO++~\cite{wang2023zero++}        & Comm. Comp.  & LLM             & PPL     & -         & -           & -         & -      & W8G8  & block    \\
		SDP4Bit~\cite{jiaxlwf0slzlt24}      & Comm. Comp.  & LLM             & PPL     & -         & -           & -         & -      & W4/G4 & -        \\
		QSDP~\cite{markovvga23}             & Comm. Comp.  & CNN/Transformer & Acc/PPL & -         & -           & -         & -      & W4/G4 & -        \\
		BinaryConnect~\cite{courbariauxbd15} & Binary       & MLP/CNN         & Acc     & binary    & -           & -         & -      & -     & -        \\
		QNN~\cite{hubaracseb17}             & Binary       & MLP/CNN         & Acc     & binary    & INT2        & INT6      & -      & -     & -        \\
		XNOR-Net~\cite{rastegariorf16}      & Binary       & CNN             & Acc     & binary    & binary      & -         & -      & -     & -        \\
		DoReFa-Net~\cite{zhou2016dorefa}    & Binary       & CNN             & Acc     & binary    & INT2        & INT6      & -      & -     & -        \\
		\mcite{Wang}{wangdmzlcccc23}        & Binary       & CNN             & Acc     & binary    & binary      & binary    & -      & -     & -        \\
		1-Bit FQT~\cite{gao20241}           & Binary       & CNN             & Acc     & binary    & binary      & binary    & -      & -     & -        \\
		\mcite{Seide}{seidefdly14}          & Comm. Comp.  & DNN             & WER     & -         & -           & -         & -      & 1-bit & -        \\
		EF-SGD~\cite{karimireddyrsj19}      & Comm. Comp.  & CNN             & Acc     & -         & -           & -         & -      & 1-bit & -        \\
		1-bit Adam~\cite{tanggarlllzh21}    & Comm. Comp.  & CNN/Transformer & Acc/F1  & -         & -           & -         & -      & 1-bit & -        \\
		0/1 Adam~\cite{lulzsh23}            & Comm. Comp.  & CNN/LLM         & Acc/PPL & -         & -           & -         & -      & 1-bit & -        \\
		1-bit LAMB~\cite{liatrh22}          & Comm. Comp.  & Transformer     & Acc/F1  & -         & -           & -         & -      & 1-bit & -        \\
		\mcite{Wu}{wuh0qwz022}              & Comm. Comp.  & CNN             & Acc     & -         & -           & -         & -      & 1-bit & -        \\
		Birder~\cite{pengqywwl23}           & Comm. Comp.  & CNN/Transformer & Acc/PPL & -         & -           & -         & -      & 1-bit & -        \\
		\bottomrule
	\end{tabular}
	\label{tab:integer_methods}
	\vspace{-0.2cm}
\end{table*}

%% file: contents/sec_fp.tex
\section{Low-Precision Floating-Point Training}
\label{sec:fp}

\input{contents/tables/tab_fp_methods}

In this section, we present low-precision training techniques based on floating-point quantization. Unlike integer quantization, converting a high-precision floating-point number to a lower-precision one involves a different process. Specifically, for converting a high-precision floating-point value (E$e$M$m$) to a lower-precision representation (E${e'}$M${m'}$), we begin by copying the lower $e'$ exponent bits from the source to the target. The mantissa is then truncated to $m'$ bits by rounding to the nearest value. To better preserve information during quantization, a scaling factor $\Delta$ is typically applied to the source value prior to conversion.
If an overflow occurs, the result is clipped directly to the maximum or minimum representable value. In the case of underflow, the value is divided by the smallest subnormal number in the low-precision format, rounded to the nearest integer, and then multiplied back by the same smallest subnormal number.
Unlike binarized training, low-precision floating-point training does not require specialized backward propagation techniques. Instead, it simply mirrors the process used in full-precision training, with the key difference being the reduced numerical precision.

Following the order from higher to lower precision, we begin with widely used 16-bit floating-point training techniques and then turn to emerging 8-bit and 4-bit approaches. Representative methods are summarized in Table~\ref{tab:fp_methods}.

Compared with integer training, floating-point methods are driven more directly by range management. As precision drops from 16-bit to 8-bit and 4-bit, progress depends less on the nominal format alone than on how scaling, accumulation, and selective high-precision exceptions are coordinated.

\subsection{16-Bit Floating-Point Training}

The adoption of 16-bit floating-point formats for deep learning training has gained widespread popularity, driven by their computational efficiency and memory savings. Modern deep learning frameworks support these formats extensively, with FP16 and BF16 emerging as the most prominent choices.

Early work by \mcite{Micikevicius}{micikeviciusnad18} established the standard mixed-precision FP16 recipe, which keeps FP32 master weights, applies loss scaling to avoid gradient underflow, and uses higher-precision accumulation for partial products. This combination made FP16 training practical without sacrificing FP32-level accuracy. Follow-up work refined this basic recipe through dynamic gradient scaling, graph-level placement to reduce casting overhead, and BF16 as a more stable 16-bit alternative because it preserves the dynamic range of FP32 \cite{zhaoval20,he00022,kalamkar2019study}.

As a result, BF16 has gradually replaced FP16 as the default format for 16-bit training. Nevertheless, with the advancement of modern GPUs, researchers are now exploring even more efficient training methods that leverage floating-point formats with lower precisions.

\subsection{Sub-8-Bit Floating-Point Training}

\textbf{Overall design.}
Early FP8 studies showed that directly lowering precision causes substantial degradation because forward and backward passes often require different numerical trade-offs. A representative solution is the hybrid-format design of HFP8, which uses E4M3 in the forward pass and E5M2 in the backward pass \cite{sunccwvsczg19}. Subsequent work broadened this direction through empirical format searches, stability analyses for large language models, communication-oriented FP8 schemes, and hardware support for exact or efficient low-bit accumulation \cite{noune20228,lee2024fp8,han2021auto,desrentes2023exact,lutz2024fused,luo2024ascend}.

As precision moved further down to FP4, the literature increasingly relied on explicit compensation mechanisms rather than the nominal format alone. Representative examples include specialized rounding and phase-splitting schemes, differentiable estimators with outlier compensation, and mixed-precision recipes that reserve FP8 for the most sensitive transformer components \cite{sunwcnacvmsg20,wang2025optimizing,zhou2025towards}. At scale, this culminates in system-level FP8 training recipes such as \mcite{Micikevicius}{micikevicius2022fp8}, FP8-LM~\cite{peng2023fp8}, and DeepSeek-V3~\cite{liu2024deepseek}, which show that sub-16-bit floating-point training can work in large models when scaling, caching, communication, and selective high-precision exceptions are co-designed.

\begin{figure}[t]
    \centering
    \includegraphics[width=\linewidth]{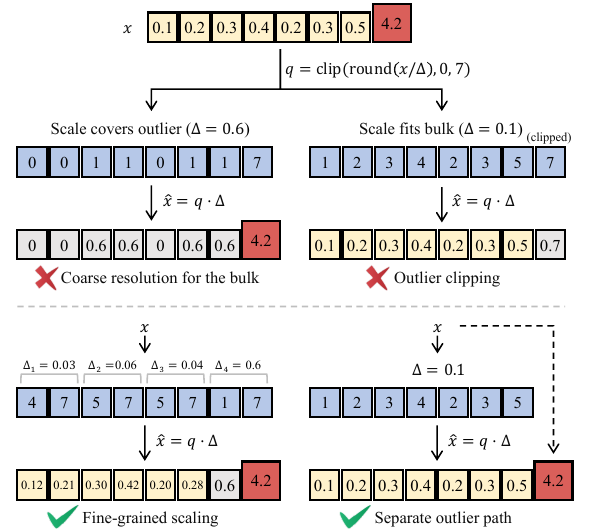}
    \caption{Illustration of the trade-off induced by a shared low-precision scale. In the toy example, choosing a large global scale preserves the outlier but yields coarse resolution for the bulk, whereas choosing a smaller scale preserves the bulk but clips the outlier. Two common remedies are finer-grained scaling and routing outliers through a separate high-precision path.}
    \label{fig:outlier}
\end{figure}

\textbf{Value scaling.}
The DeepSeek-V3 example above reflects a broader pattern in low-precision floating-point training. Once precision drops to FP8 or FP4, successful recipes depend critically on how values are scaled before quantization. Earlier studies therefore devoted substantial effort to designing scaling strategies that balance dynamic range against effective resolution.
Figure~\ref{fig:outlier} summarizes the core trade-off behind this line of work. With one shared low-precision scale, covering an outlier wastes resolution on the bulk of the tensor, whereas fitting the bulk clips the outlier. Much of the subsequent literature can be viewed as different ways of relaxing this trade-off through finer-grained scaling or explicit outlier handling.

Against this backdrop, existing methods differ mainly in how broadly the scale is shared, how dynamically it is updated, and whether outliers are handled implicitly or explicitly. Early work mainly focused on loss or gradient scaling, either by adapting the update frequency of a global scale or by learning per-layer gradient scales during training \cite{mellempudi2019mixed,sunwcnacvmsg20}. Later methods shifted toward more automated and finer-grained scale control, including tensor-wise rescaling, format search for gradient scaling, initialization-based unit scaling, global coordination across distributed GPUs, dynamic per-tensor scaling for LLMs, and explicit scale propagation through computational graphs \cite{cambierbgent20,chmielbshbs21,blake2023unit,peng2023fp8,perez2023training,balancca2024scalify}.

\textbf{Result accumulation.}
Another recurring issue is that low-precision inputs do not automatically imply low-precision-safe accumulation. Representative work therefore studies how to preserve gradient fidelity through chunk-based accumulation, stochastic rounding, reduced-precision master weights, statistical analyses of the precision needed for long accumulations, and dedicated FP8 MAC designs \cite{mellempudi2019mixed,wangcbcg18,sakrwccasg19,alifs24}. The shared lesson is that accumulation precision often becomes the real bottleneck once input formats are pushed to FP8 or FP4.

\textbf{Optimizer states.}
Compared with activations and weights, optimizer states were incorporated later into floating-point compression recipes. Initial efforts showed that mixed FP8/FP16 optimizer designs can substantially reduce memory while preserving stability \cite{peng2023fp8,fishman2024scaling}. More recent work argues that naive FP8 state quantization under-utilizes the available range, motivating co-designed solutions that expand optimizer-state ranges before quantization and jointly optimize state and activation compression granularity \cite{xi2024coat}.

\textbf{Block floating-point.}
Besides standard low-bit floating point, block floating point (BFP) offers an alternative that shares exponents across blocks while retaining much of floating point's dynamic-range advantage. Representative work includes HBFP, which uses BFP for dot products while keeping other operations in standard floating point, and later studies showing that relatively short mantissas can still recover FP32-level accuracy when the format is used consistently \cite{drumondljf18,harma2022accuracy}.

Overall, sub-8-bit floating-point training has evolved from format exploration to full-stack co-design. The frontier is no longer whether FP8 or FP4 arithmetic can be used at all, but how much of the pipeline can be pushed down while retaining fine-grained scaling, protected states, and efficient kernels.

%% file: contents/tables/tab_fp_methods.tex
\begin{table*}[t]
	\centering
	\caption{Representative floating-point training methods.}
	\setlength{\tabcolsep}{2.4pt}
	\begin{tabular}{llllccccccc}
		\toprule
		Paper                                     & Family       & Model               & Metrics & Weights  & Activations & Gradients & Optim. & Comm. & Grouping     \\
		\midrule
		\mcite{Micikevicius}{micikeviciusnad18}   & FP           & CNN/RNN/GAN         & Acc/BLEU/PPL     & FP16     & FP16        & FP16      & -      & -     & -            \\
		\mcite{Kalamkar}{kalamkar2019study}       & FP           & CNN/RNN/GAN/RecSys  & Acc/WER/PPL/AUC  & BF16     & BF16        & BF16      & -      & -     & -            \\
		\mcite{Wang}{wangcbcg18}                  & FP           & CNN                 & Acc              & FP8      & FP8         & FP8       & -      & -     & -            \\
		\mcite{Mellempudi}{mellempudi2019mixed}   & FP           & CNN/RNN/Transformer & Acc/BLEU         & FP8      & FP8         & FP8       & -      & -     & -            \\
		HFP8~\cite{sunccwvsczg19}                 & FP           & CNN/RNN/Transformer & Acc/mAP/BLEU/WER & FP8      & FP8         & FP8       & -      & -     & -            \\
		DeepSeek-V3~\cite{liu2024deepseek}        & FP           & LLM                 & Acc/PPL          & FP8      & FP8         & FP8       & BF16   & FP8   & block/tile   \\
		HBFP~\cite{drumondljf18}                  & FP           & CNN/RNN/Transformer & Acc/BLEU/PPL     & BFP      & BFP         & -         & -      & -     & block        \\
		\mcite{Zhao}{zhaoval20}                   & FP           & CNN                 & Acc              & FP16     & FP16        & FP16      & -      & -     & layer        \\
		\mcite{Chmiel}{chmielbshbs21}             & FP           & CNN                 & Acc              & -        & -           & FP6       & -      & -     & -            \\
		\mcite{Noune}{noune20228}                 & FP           & CNN/Transformer     & Acc/BLEU         & FP8      & FP8         & FP8       & -      & -     & -            \\
		\mcite{Micikevicius}{micikevicius2022fp8} & FP           & CNN/RNN/LLM         & Acc/PPL          & FP8      & FP8         & FP8       & -      & -     & tensor       \\
		S2FP8~\cite{cambierbgent20}               & FP           & CNN/Transformer     & Acc/BLEU         & FP8      & FP8         & FP8       & -      & -     & tensor       \\
		\mcite{Blake}{blake2023unit}              & FP           & LLM                 & Acc/PPL          & FP16     & FP16        & FP16      & -      & -     & -            \\
		\mcite{Perez}{perez2023training}          & FP           & LLM                 & Acc/PPL          & FP8      & FP8         & FP8       & -      & -     & tensor       \\
		Scalify~\cite{balancca2024scalify}        & FP           & LLM                 & PPL              & FP8      & FP8         & FP8       & FP16   & -     & -            \\
		\mcite{Luo}{luo2024ascend}                & FP           & CNN/LLM             & Acc/PPL          & HiFloat8 & HiFloat8    & HiFloat8  & -      & -     & tensor       \\
		GradScale~\cite{sunwcnacvmsg20}           & FP           & CNN/RNN/Transformer & Acc/BLEU/PPL/WER & FP4      & FP4/INT4    & FP4       & -      & -     & layer        \\
		\mcite{Wang}{wang2025optimizing}          & FP           & LLM                 & PPL              & FP4      & FP4         & -         & -      & -     & vector       \\
		\mcite{Zhou}{zhou2025towards}             & FP           & LLM                 & Acc/PPL          & FP4      & FP4         & FP4       & -      & -     & block        \\
		FP8-LM~\cite{peng2023fp8}                 & Optim. Comp. & LLM                 & Acc/PPL          & FP8      & FP8         & FP8       & FP8/16 & FP8   & tensor       \\
		\mcite{Fishman}{fishman2024scaling}       & Optim. Comp. & LLM                 & Acc/PPL          & FP8      & FP8         & FP8       & FP8    & -     & -            \\
		COAT~\cite{xi2024coat}                    & Act. Comp.   & LLM/VLM             & Acc/PPL          & -        & FP8         & -         & FP8    & -     & tensor/group \\
		APS~\cite{han2021auto}                    & Comm. Comp.  & CNN/RNN/Transformer & Acc              & -        & -           & -         & -      & FP8   & -            \\
		\bottomrule
	\end{tabular}
	\label{tab:fp_methods}
\end{table*}

%% file: contents/sec_custom.tex
\section{Custom Numerical Formats}
\label{sec:custom}

\input{contents/tables/tab_custom_methods}

Besides commonly used fixed-point and floating-point representations, some works propose customized numerical formats for low-precision training. It is important to note that certain customized formats, such as NormalFloat~\cite{dettmersphz23}, are specifically designed for pretrained fixed parameters, which only participate in the forward inference stage during training. As such, we exclude these works from consideration, as they do not directly contribute to low-precision training. Although the encodings differ, most of these formats explore the same design space by determining how much exponent information should be shared, where scaling should be attached, and how hardware simplicity should be traded against adaptability to nonuniform tensor statistics. Representative methods are summarized in Table~\ref{tab:custom_methods}.

Custom numerical formats can largely be understood as alternative operating points between fixed-point and floating-point design. Posit reallocates bits dynamically among sign, regime, exponent, and fraction fields to improve dynamic range and accuracy under a fixed bit budget~\cite{gustafsony17,de2019posits,lufxlw21}. Flexpoint instead shares a dynamically adjusted exponent across each tensor, preserving much of fixed-point efficiency while adapting to changing ranges \cite{kosterwwnbcehhk17}. FloatSD modifies the value representation itself through signed-digit encoding to simplify multiply-heavy computation, whereas MLS combines element-wise and group-wise scaling to balance representational flexibility with efficient low-overhead accumulation~\cite{linskc19,zhongndzzzwy22}. More recent microscaling methods push this idea further by assigning shared scaling factors to small blocks rather than whole tensors, which better matches local value distributions and has enabled near-lossless FP4-style training when paired with suitable rounding, transforms, and stabilization strategies~\cite{rouhani2023microscaling,tseng2025training,chen2025oscillation,hu2025elucidating}.

Overall, custom formats are best viewed as targeted operating points between fixed-point and floating-point extremes rather than a disconnected set of tricks. Their strongest results appear when the encoding is paired with matching scaling rules and hardware or dataflow assumptions, as illustrated most clearly by recent microscaling-based FP4 training.

%% file: contents/tables/tab_custom_methods.tex
\begin{table*}[t]
	\centering
	\caption{Representative custom-format training methods.}
	\setlength{\tabcolsep}{8pt}
	\begin{tabular}{lllllcccccc}
		\toprule
		Paper                                    & Family & Model   & Metrics    & Weights      & Activations  & Gradients    & Optim. & Comm. & Grouping \\
		\midrule
		\mcite{Lu}{lufxlw21}                     & Custom & DNN     & Acc/PPL    & posit(8,1)   & posit(8,1)   & posit(8,1)   & -      & -     & tensor   \\
		FloatSD~\cite{linskc19}                  & Custom & CNN     & Acc        & FloatSD      & FP8          & FP8          & -      & -     & -        \\
		Flexpoint~\cite{kosterwwnbcehhk17}       & Custom & CNN/GAN & Acc/FID    & flex16+5     & flex16+5     & flex16+5     & -      & -     & tensor   \\
		MLS~\cite{zhongndzzzwy22}                & Custom & CNN     & Acc/Energy & MLS⟨2,1⟩     & MLS⟨2,1⟩     & MLS⟨2,1⟩     & -      & -     & group    \\
		\mcite{Rouhani}{rouhani2023microscaling} & Custom & LLM     & Acc/PPL    & MXFP6        & MXFP6        & MXFP6        & -      & -     & block    \\
		\mcite{Tseng}{tseng2025training}         & Custom & LLM     & PPL        & MXFP4        & MXFP4        & MXFP4        & -      & -     & block    \\
		\mcite{Chen}{chen2025oscillation}        & Custom & ViT     & Acc        & MXFP4        & MXFP4        & MXFP4        & -      & -     & block    \\
		\bottomrule
	\end{tabular}
	\label{tab:custom_methods}
\end{table*}

%% file: contents/sec_cross.tex
\section{Cross-Cutting Perspectives and Insights}
\label{sec:cross}

\input{fig/forest_cross}

After reviewing low-precision training methods by numerical format, we revisit the literature from two complementary cross-cutting perspectives. We first summarize the shared numerical bottlenecks that recur across different format families, and then discuss how these bottlenecks relate to robustness and model reliability in practical LLM pipelines.

\subsection{A Problem-Centric View Across Formats}

The format-driven taxonomy is useful for locating methods by representation and hardware affinity. Yet many methods using different formats address the same numerical bottlenecks. A complementary problem-centric view therefore helps reveal common structure across fixed-point, integer, floating-point, and customized formats. In most cases, gains do not come from a datatype change alone, but from coordinating range management, update-fidelity protection, precision allocation, and system/dataflow co-design. This perspective explains why methods built on different encodings often converge to similar design choices.

Figure~\ref{fig:cross} summarizes the recurring issues shared across the literature and the representative remedies developed for them. One branch concerns \emph{dynamic range mismatch and outlier handling}, where a few large activations, tiny gradients, or block-wise distribution shifts can induce overflow, underflow, or excessive clipping after precision reduction. A second branch concerns \emph{accumulation and state fidelity}, since repeated rounding may erase small but important updates or distort optimizer statistics even when forward values remain representable. A third branch captures \emph{precision allocation}, reflecting the fact that layers, operators, and training stages exhibit markedly different sensitivity to low precision. A fourth branch concerns \emph{execution and communication}, because arithmetic compression alone brings limited benefit when cast overhead, activation storage, or distributed synchronization still dominate runtime and memory.

These issues are tightly coupled. Better scaling reduces accumulator pressure, protected updates make aggressive precision schedules more viable, and system support determines whether fine-grained scaling or selective high-precision exceptions remain affordable in practice. This interdependence explains why methods from different format families often arrive at analogous solutions.

The practical implication is that moving from 16-bit training to 8-bit or 4-bit training usually requires simultaneous progress along several branches in Figure~\ref{fig:cross}. FP16/BF16 training became robust only after range management and update protection were established. Recent FP8, FP4, and MXFP4 results push further by combining finer-grained scaling, protected optimizer states or accumulators, selective high-precision exceptions, and cast-aware kernels or communication compression. New numerical formats therefore shift the operating point, whereas stability and efficiency still depend on how well these shared design levers are coordinated. This trend suggests that future progress is likely to come less from a single universally optimal format than from better coordination between format choice, scaling granularity, optimizer design, and system support.

\input{contents/tables/tab_family_compare}

\input{contents/tables/tab_controlled_benchmark}

A comparable optimization paradigm extends to broader LLM-empowered applications. Because many VLMs inherit the underlying transformer backbone and distributed training infrastructure, the previously identified precision strategies remain highly relevant in multimodal contexts. For instance, \mcite{Wortsman}{wortsmandzmfs23} stabilize large-scale VLM training by keeping sensitive updates in 16-bit while quantizing the forward pass and input gradients to 8-bit integers, reflecting the same update-fidelity principle used in LLM training. Meanwhile, MM1~\cite{mckinzie2024mm1} shows that multimodal quality is especially sensitive to factors such as the image encoder, image resolution, and image token count, which suggests that these levers should be applied in a component-aware rather than uniform way. Low-precision recipes are therefore not confined to text-only LLMs, but their extension to VLMs and related empowered applications requires modality-aware precision allocation and evaluation.

\subsection{Performance Trade-offs Across Families}

Directly comparing the empirical results of published low-precision training methods is confounded by heavily misaligned experimental setups across the literature, encompassing disparate models, datasets, hardware platforms, and stabilization recipes. Because quantitative metrics cannot be strictly aligned, we first synthesize existing results into a coarse-grained categorical comparison that captures the typical operating region of each method family. To complement this broad synthesis with a more fine-grained analysis, we subsequently conduct a controlled benchmark under a strictly unified setup.

Table~\ref{tab:family_compare} presents this qualitative synthesis. By abstracting away misaligned experimental details, we derive broad performance tiers rather than a potentially misleading direct ranking. This macroscopic view reveals a consistent paradigm. Integer and binary methods maximize compression along the primary training path. Reduced floating-point and custom formats offer the most robust balance between efficiency and numerical stability. Meanwhile, specialized techniques that compress optimizer states, activations, or communications serve as bottleneck-specific optimizations rather than complete alternatives to full low-precision numerical formats.

To ground the broad family-level tiers with a precise empirical comparison, Table~\ref{tab:controlled_benchmark} reports a unified DistilBERT~\cite{sanh2019distilbert} fine-tuning benchmark on AG News~\cite{zhang2015character} and IMDB~\cite{maas2011learning}. Both tasks use a shared AdamW-based single-seed protocol, with AG News trained on a 50k-sample subset for 5 epochs and IMDB for 4 epochs. Native FP32, BF16, FP16, and FP8 are evaluated on an NVIDIA RTX PRO 6000 GPU, while simulated FP4 and INT8 are included only to assess accuracy degradation at extreme low precisions. The setup fixes the model, task family, and training protocol, but it intentionally remains much smaller than foundation-model pre-training or alignment workloads.

Within this limited DistilBERT fine-tuning setting, BF16, FP16, and FP8 achieve accuracy close to FP32 on both AG News and IMDB, whereas INT8 and especially FP4 show larger degradation. These results suggest that sufficient numerical range remains important even in compact-model fine-tuning, while aggressive precision reduction requires stronger stabilization recipes before it can be generalized to larger or more demanding training regimes. The modest FP8 gains in throughput and memory also reflect practical overheads from higher-precision optimizer states, auxiliary tensors, casting, normalization, and memory movement.

Importantly, this benchmark is a lightweight diagnostic, not a substitute for foundation-model-scale pre-training or instruction-tuning evaluation. Its scope is limited to a compact encoder, two text-classification datasets, one hardware platform, and simulated extreme-low-precision formats without corresponding efficiency measurements. It complements the survey-level synthesis by showing how representative formats behave under one controlled fine-tuning setup.

\subsection{Robustness and Model Reliability}

While efficiency metrics (\eg, memory footprint, throughput, and accuracy) are critical, they insufficiently capture the reliability of low-precision recipes in modern foundation-model pipelines. Across representative low-precision papers, evaluation is still dominated by loss or perplexity, downstream task accuracy, memory, and throughput, leaving a much thinner evidence base on reasoning, alignment, and execution stability. Table~\ref{tab:robustness_dimensions} summarizes several robustness-related dimensions that can complement conventional efficiency-oriented evaluations.

\begin{table}[t]
	\centering
	\caption{Robustness-related evaluation dimensions that are often underrepresented in low-precision training studies.}
	\label{tab:robustness_dimensions}
	\setlength{\tabcolsep}{2pt}
    \begin{tabular}{
            >{\raggedright\arraybackslash}p{0.15\linewidth}
            >{\raggedright\arraybackslash}p{0.41\linewidth}
            >{\raggedright\arraybackslash}p{0.38\linewidth}
        }
        \toprule
        Dimension & Evaluation focus & Low-precision concern \\
        \midrule
        Capability robustness
        & reasoning; math; coding; long-context tasks
        & fragile capability loss hidden by average accuracy \\
        \addlinespace
        Alignment robustness
        & refusal consistency; preference alignment; safety behavior
        & safety-relevant behavior shift despite stable perplexity \\
        \addlinespace
        Tool-use reliability
        & API selection; JSON validity; argument exactness
        & local token errors breaking structured execution \\
        \bottomrule
    \end{tabular}
\end{table}

Recent results nevertheless point to concrete reliability risks rather than a purely hypothetical concern. DeepSeek-V3 shows that stable FP8 training still requires sensitive modules and update paths to remain in higher precision~\cite{liu2024deepseek}. Beyond optimization instability, low-precision artifacts can disproportionately degrade complex cognitive tasks long before headline perplexity or general language quality collapses~\cite{li2025quantization}. More broadly, reasoning robustness benchmarks show that LLM reasoning can be sensitive to instruction changes, numerical perturbations, and missing information, while low-bit mathematical-reasoning analyses further locate degradation in step-level reasoning failures~\cite{yu2025benchmarking,li2025quantization}. These observations suggest that aggregate loss or average accuracy can overestimate the true capability retained under aggressive precision reduction.

A second critical concern is whether precision recipes validated during pre-training remain reliable during post-training alignment. Existing evidence is still limited but not absent. Frameworks like \mcite{Dettmers}{dettmersphz23} and \mcite{Peng}{peng2023fp8} demonstrate the feasibility of instruction tuning and RLHF on low-precision backbones, yet they do not provide broad stress tests of safety-sensitive behaviors. Recent trustworthiness and safety-focused evaluations of compressed or quantized LLMs further suggest that compression choices can alter safety, reliability, fairness, and human-evaluated behavior across benchmarks~\cite{hong2024decoding,kharinaev2026investigating,chen2025assessing}. Low-precision feasibility during SFT, DPO, or RLHF should therefore not be conflated with verified alignment robustness~\cite{thakkar2024deep}.

Moreover, tool-augmented LLMs impose a zero-tolerance policy for localized token errors. As formalized by benchmarks like API-Bank~\cite{li2023api} and ToolLLM~\cite{qintoolllm}, successful tool invocation relies on accurate API retrieval and exact JSON argument formatting. Yet tool-use proficiency is largely overlooked in low-precision training evaluations. While a minor quantization-induced deviation may be harmless in free-form text, it can irreparably corrupt a structured payload and crash an execution pipeline. This remains a clear gap between current low-precision evaluation practice and deployment-oriented reliability requirements.

%% file: fig/forest_cross.tex
\tikzstyle{leaf}=[
    align=left,
    inner xsep=10pt,
    inner ysep=3pt,
]

\begin{figure*}[htpb]
    \centering
    \begin{forest}
        forked edges,
        for tree={
            font=\sffamily\bfseries,
            text=textcolor,
            grow=east,
            reversed=true,
            anchor=base west,
            parent anchor=east,
            child anchor=west,
            base=center,
            rectangle,
            draw=black,
            rounded corners,
            align=left,
            text centered,
            minimum width=4em,
            edge+={black, line width=1pt},
            s sep=3pt,
            inner xsep=2pt,
            inner ysep=3pt,
            line width=0.8pt,
            ver/.style={rotate=90, child anchor=north, parent anchor=south, anchor=center},
            myline/.style={line width=0.8pt},
        },
        where level=0{fill=rootcolor,text=white,text width=10em,font=\footnotesize,}{},
        where level=1{fill=level1color,text=white,text width=12em,font=\footnotesize,}{},
        where level=2{fill=level3color,text width=32em,font=\scriptsize,leaf,}{},
        [
        {Cross-Cutting View}, ver,
            [
            {Range Mismatch and Outliers}, name=range_root
                [
                {
                \textbf{Range control:} \mcite{Micikevicius}{micikeviciusnad18}, \mcite{Zhao}{zhaoval20}, FP8-LM~\cite{peng2023fp8}, \mcite{Perez}{perez2023training}, Scalify~\cite{balancca2024scalify}\\
                \textbf{Fine-grained scaling:} DeepSeek-V3~\cite{liu2024deepseek}\\
                \textbf{Outlier shaping:} \mcite{Xi}{xilcz23}, HLQ~\cite{kim2024hlq}, \mcite{Rouhani}{rouhani2023microscaling}, \mcite{Tseng}{tseng2025training}, \mcite{Chen}{chen2025oscillation}
                }
                ,
                name=range_leaf,
                edge={draw=none}
                ]
            ]
            [
            {Accumulation and State Fidelity}, name=state_root
                [
                {
                \textbf{Update protection:} \mcite{Micikevicius}{micikeviciusnad18}, \mcite{Park}{park2018training}, \mcite{Sakr}{sakrwccasg19}\\
                \textbf{Low-bit optimizer states:} \mcite{Dettmers}{dettmerslsz22}, \mcite{Li}{licz23}, \mcite{Fishman}{fishman2024scaling}\\
                \textbf{State range expansion:} COAT~\cite{xi2024coat}
                }
                ,
                name=state_leaf,
                edge={draw=none}
                ]
            ]
            [
            {Precision Allocation}, name=precision_root
                [
                {
                \textbf{Temporal scheduling:} FracTrain~\cite{fuyz0lgwl20}, MuPPET~\cite{rajagopalvvb20}, CPT~\cite{fuglydcl21}\\
                \textbf{Forward/backward decoupling:} HFP8~\cite{sunccwvsczg19}\\
                \textbf{Operator-aware exceptions:} \mcite{Zhou}{zhou2025towards}, DeepSeek-V3~\cite{liu2024deepseek}
                }
                ,
                name=precision_leaf,
                edge={draw=none}
                ]
            ]
            [
            {Execution and Communication}, name=system_root
                [
                {
                \textbf{Cast-aware execution:} Campo~\cite{he00022}\\
                \textbf{Low-bit dataflow and memory:} Jetfire~\cite{xicztcz24}, ActNN~\cite{chenzywsm021}, GACT~\cite{liuzwcchcltgmc22}, COAT~\cite{xi2024coat}\\
                \textbf{Distributed compression:} FP8-LM~\cite{peng2023fp8}, ZeRO++~\cite{wang2023zero++}, QSDP~\cite{markovvga23}, SDP4Bit~\cite{jiaxlwf0slzlt24}
                }
                ,
                name=system_leaf,
                edge={draw=none}
                ]
            ]
        ]
        \draw[line width=1pt] (range_root) -- (range_leaf);
        \draw[line width=1pt] (state_root) -- (state_leaf);
        \draw[line width=1pt] (precision_root) -- (precision_leaf);
        \draw[line width=1pt] (system_root) -- (system_leaf);
    \end{forest}
    \caption{A cross-cutting view of recurring issues and representative solutions in low-precision training.}
    \label{fig:cross}
    \vspace{-0.3cm}
\end{figure*}

%% file: contents/tables/tab_family_compare.tex
\begin{table*}[t]
	\centering
	\caption{Family-level comparison of representative low-precision training strategies. Relative benefits are denoted categorically as H (High), M (Medium), and L (Low). Ratings reflect the typical degree of reduction in training memory, the impact on the main compute or communication bottlenecks, and the ability to remain close to high-precision baselines under representative recipes; interval ratings indicate the typical variance within a family.}
	\setlength{\tabcolsep}{3pt}
	\begin{tabular}{lllccc}
		\toprule
		Family      & Representative formats / methods       & Typical target                               & Memory efficiency & Speed benefit & Accuracy retention \\
		\midrule
		FX          & fixed16, dynamic fixed-point           & weights, activations, gradients              & M                 & M             & M-H                \\
		INT         & INT8, INT4                             & weights, activations, gradients              & H                 & M-H           & M                  \\
		FP          & FP16, BF16, FP8, FP4                   & weights, activations, gradients              & M-H               & M-H           & M-H                \\
		Custom      & posit, Flexpoint, FloatSD, MLS, MX     & custom low-precision tensors                 & M                 & M-H           & M-H                \\
		Binary      & 0/1, 1/-1                              & weights, activations, gradients              & H                 & H             & L                  \\
		Opt. comp.  & 8-bit Optimizer, 4-bit Shampoo         & moments, momentum, preconditioners           & H                 & M             & H                  \\
		Act. comp.  & ActNN, GACT, COAT                      & stored activations, backward signals         & H                 & M             & H                  \\
		Comm. comp. & QSGD, ZeRO++, 1-bit Adam   & communicated gradients or optimizer states & L                 & H             & H                  \\
		\bottomrule
	\end{tabular}
	\label{tab:family_compare}
\end{table*}

%% file: contents/tables/tab_controlled_benchmark.tex
\begin{table}[t]
	\centering
	\caption{Controlled benchmarking of DistilBERT fine-tuning on AG News and IMDB. Empirical throughput and peak memory are reported for native hardware formats, whereas simulated FP4 and INT8 are included solely to evaluate accuracy degradation at extreme low precisions without corresponding hardware efficiency metrics. Higher is better for accuracy and speed, whereas lower is better for memory.}
	\label{tab:controlled_benchmark}
	\setlength{\tabcolsep}{3pt}
	\begin{tabular}{ll|ccc|ccc}
		\toprule
		\multirow{3}{*}{Recipe} & \multirow{3}{*}{Impl.} 
		& \multicolumn{3}{c|}{AG News} 
		& \multicolumn{3}{c}{IMDB} \\
		\cmidrule(lr){3-5} \cmidrule(lr){6-8}
		& 
		& Acc. & Speed & Memory 
		& Acc. & Speed & Memory \\
		& 
		& (\%) & (data/s) & (GB) 
		& (\%) & (data/s) & (GB) \\
		\midrule
		FP32        & Native & 93.45 & 1263.3 & 3.28 & 91.18 & 476.2  & 2.06 \\
		BF16        & Native & 93.49 & 2944.8 & 2.46 & 91.47 & 1005.3 & 1.71 \\
		FP16        & Native & 93.41 & 2998.0 & 2.46 & 91.50 & 1030.5 & 1.71 \\
		FP8         & Native & 93.58 & 3205.0 & 2.24 & 91.38 & 928.7  & 1.58 \\
		FP4         & Sim.   & 89.80 & --     & --   & 80.44 & --     & --   \\
		INT8        & Sim.   & 91.36 & --     & --   & 89.19 & --     & --   \\
		\bottomrule
	\end{tabular}
\end{table}

%% file: contents/sec_qat.tex
\section{Quantization-Aware Training Techniques}
\label{sec:qat}

In addition to low-precision training, the deployment of LLMs with low precision during inference has also been extensively studied. While low-precision training aims to reduce the precision of both the forward and backward passes during the training process, low-precision inference focuses solely on reducing the precision of weights and activations during the forward pass. This typically involves the application of quantization techniques.

Works in this domain can be broadly divided into Post-Training Quantization (PTQ) and QAT. PTQ methods typically quantize a pre-trained model without further training and rely on heuristics or optimization strategies to adapt the model to lower precision. In contrast, QAT incorporates quantization into the training process, often via fake-quantization operators or learned quantization parameters, allowing the model to adapt to deployment-time low-precision arithmetic during optimization~\cite{jacob2018quantization,krishnamoorthi2018quantizing,esser2020learned}. Although there is a large body of work on PTQ techniques for LLMs~\cite{wei2022outlier,xiao2023smoothquant,yang2025raana,lin2024awq,zhao2024atom,liu2025septq,zhuz00l0t23,li2023fp8}, these methods are beyond the scope of this survey, and thus will not be covered here. QAT methods share some similarities with low-precision training, so the remainder of this section briefly reviews representative QAT studies.

\textbf{Binary and ternary QAT.}
Extreme low-bit QAT for LLMs is represented by the BitNet line of work. BitNet~\cite{wang2023bitnet} established the feasibility of 1-bit Transformers by combining binary weights, 8-bit activations, and STE-based training while retaining higher-precision gradients. BitNet b1.58~\cite{ma2024era} extended this idea from binary to ternary parameters through absmean quantization:
\begin{equation}
    W_q = \max\left(-1, \min\left(1, \mathrm{round}\left(\frac{W}{\gamma + \epsilon}\right)\right)\right),
\end{equation}
where $\gamma=\frac{1}{nm}\sum_{ij} |W_{ij}|$ indicates absolute value of all elements in $w$. More recent work further improves stability through progressive schedules that transition from higher-precision pretraining to 1.58-bit quantization, reducing the disruption caused by immediate extreme quantization~\cite{nielsen2025continual}.

\textbf{QAT with KD.}
Another line of work combines QAT with knowledge distillation so that a low-bit student can better match a stronger full-precision or pretrained teacher. Representative examples include data-free distillation for 4-bit LLaMA quantization and teacher-guided sub-4-bit training with quantization-aware clipping and asymmetric quantizers~\cite{liuo0csmskc24,duzcgccx24}.

\textbf{Other QAT methods.}
Beyond these representative directions, recent work has focused on improving the efficiency and stability of QAT itself. This includes reducing QAT memory overhead through staged optimization of model and quantization parameters, improving 1-bit training with Hadamard normalization and more reliable gradient estimation, and explaining QAT instability through loss-landscape sharpness, which motivates feature perturbation as an implicit regularizer~\cite{chen2024efficientqat,panferov2025quest,pang2025stabilizing}.

%% file: contents/sec_system.tex
\section{System Support for Low-Precision Training}
\label{sec:system}

\begin{table*}[htbp]
	\renewcommand\tabcolsep{5.0pt}
	\centering
	\caption{Summary of Software Frameworks and Libraries for Low-Precision Training Support}
	\begin{tabular}{lll}
		\toprule
		Framework/Library  & Key Low-Precision Features                                          & Primary Focus / Notes                             \\
		\midrule
		PyTorch            & Native AMP, GradScaler                                              & General-purpose deep learning                     \\
		TensorFlow         & Native Mixed Precision, Auto loss scaling                           & General-purpose deep learning                     \\
		JAX                & Explicit precision control                                          & General-purpose deep learning                     \\
		PaddlePaddle       & AMP with automatic operator casting (FP16 focus)                    & General-purpose deep learning                     \\ \addlinespace 
		DeepSpeed          & Enhanced AMP, ZeRO integration, Exp. 4/8-bit QAT, 1-bit optimizers  & Large-scale model training (extends PyTorch)      \\
		Megatron-LM/Core   & Optimized kernels, FP8 support (Hopper), Parallelism APIs           & LLM training (PyTorch-based)                      \\ \addlinespace
		bitsandbytes       & 8-bit optimizers, 4/8-bit quantization functions                    & Lightweight CUDA wrappers, LLM memory reduction   \\
		TorchAO            & PyTorch-native float8/QAT/quantization workflows                    & Training-to-serving model optimization            \\
		MS-AMP             & Automatic mixed precision with FP8 support                          & Large-scale mixed-precision training              \\
		Colossal-AI        & Simplified AMP (FP16/BF16 planned), Exp. FP8 linear/communication   & Simplifying large model training                  \\
		Transformer Engine & FP8 Transformer layers and scaling utilities                        & Optimized FP8 Transformer kernels                 \\
		NVIDIA Apex        & Early AMP \& distributed tools                                      & Historical PyTorch extension                      \\
		\bottomrule
	\end{tabular}
	\label{tab:framework}
\end{table*}

Low-precision training techniques offer significant benefits for model training, primarily by reducing memory footprint, decreasing memory bandwidth requirements, and potentially accelerating computations on compatible hardware. However, effectively leveraging these advantages requires sophisticated system-level support. Software frameworks and specialized libraries play a pivotal role in managing the complexities of low-precision formats, such as ensuring numerical stability, performing efficient data type conversions, and optimizing hardware utilization. This section provides an overview of the software frameworks and libraries that offer essential system support for enabling and facilitating low-precision DNN training. Table~\ref{tab:framework} provides a summary of introduced frameworks and libraries.

\textbf{Mainstream frameworks.}
Several widely adopted deep learning frameworks support low-precision training through built-in tools and abstractions.
PyTorch~\cite{paszke2019pytorch}, known for its flexible design, includes native Automatic Mixed Precision (AMP), which automatically casts operations between FP32 and lower-precision formats (typically FP16 or BF16). AMP handles gradient scaling internally to preserve numerical stability, requiring minimal user intervention.
TensorFlow~\cite{tensorflow2015} provides similar mixed-precision capabilities, automatically applying loss scaling and optimization routines for efficient execution on accelerators like GPUs and TPUs.
JAX~\cite{jax2018github}, a high-performance numerical library, allows precise control over data types, enabling researchers to experiment with precision settings at the operation or module level.
PaddlePaddle~\cite{ma2019paddlepaddle} integrates AMP functionality as well, enabling efficient FP16 computation while preserving critical operations in FP32 to maintain accuracy.

\textbf{Frameworks for LLMs.}
Training large-scale models, especially transformers, requires frameworks that blend low-precision arithmetic with memory and parallelization optimizations.
DeepSpeed~\cite{rasley2020deepspeed} extends PyTorch for efficient large-model training, combining memory-saving strategies like ZeRO~\cite{rajbhandari2020zero} with optimized mixed-precision support. It also supports quantization-aware optimizers and low-precision (4-bit, 8-bit) training, including gradient compression for scalable distributed training.
Megatron-LM, now formalized as Megatron-Core, is a modular PyTorch-based framework for LLM training. It incorporates tensor, pipeline, and sequence parallelism, with support for FP8 precision on hardware such as NVIDIA Hopper GPUs.

\textbf{Specialized libraries and extensions.}
A growing ecosystem of libraries offers targeted support for low-precision computation, quantized inference, and hardware acceleration.
bitsandbytes provides memory-efficient 8-bit optimizers~\cite{dettmers20228bit} and matrix multiplication kernels~\cite{dettmers2022llm}, supporting both inference and training.
TorchAO~\cite{or2025torchao} extends the PyTorch ecosystem with native float8 training, quantization-aware training, and low-bit quantization workflows, helping bridge research prototypes and deployment-oriented optimization.
MS-AMP~\cite{peng2023fp8} is a Microsoft mixed-precision library that emphasizes automated FP8-enabled training workflows for large-scale models.
Colossal-AI~\cite{li2023colossal} simplifies the use of advanced training strategies, including AMP and experimental FP8 features, along with FP8-based communication compression for efficient distributed training.
Transformer Engine focuses on FP8 transformer kernels and scaling utilities on recent NVIDIA GPUs, while Apex~\cite{apex} mainly remains useful as a historical precursor to native PyTorch AMP.

\textbf{Hardware-platform portability.}
The practical deployment of low-precision training also depends on whether the target accelerator exposes matching datatype, kernel, compiler, and communication support. Although many early large-scale recipes were developed in CUDA-centered environments, AMD GPUs and the ROCm ecosystem now provide an important non-NVIDIA training path. ROCm supports PyTorch-oriented training workflows on AMD Instinct accelerators, and ROCm's precision documentation reports matrix-core support for FP8 E4M3/E5M2 on CDNA3-class GPUs. ROCm TransformerEngine further provides transformer-oriented FP8 components and GEMM tuning paths for AMD platforms~\cite{rocmTransformerEngine}. These capabilities make BF16-, FP16-, INT8-, and FP8-oriented recipes relevant beyond NVIDIA hardware. At the same time, portability is not automatic. Methods relying on CUDA-specific custom kernels, NVIDIA Transformer Engine APIs, native FP4/NVFP4 execution, or a particular Tensor Core dataflow may require ROCm-specific kernel implementations, format conversion, and renewed empirical validation. Thus, practical low-precision training should be treated as a hardware-software co-design problem rather than only a numerical-format choice.

%% file: contents/sec_future.tex
\section{Future Directions}
\label{sec:future}

Despite impressive progress, low-precision training of foundation models still faces several open challenges. Addressing these issues is crucial for further scaling while preserving both performance and training stability. Below, we outline key challenges along with promising directions for future research.

\textbf{Advanced quantization methods.}
Linear quantization is widely used due to its simplicity and hardware compatibility but struggles to capture the diverse statistics of weights, activations, and gradients. Future work should explore non-linear methods like logarithmic or learned quantization, which offer better dynamic range and adaptability. These approaches are promising for low-precision training, especially when linear quantization falls short.

\textbf{Ultra low-precision training.}
While 8-bit precision performs well, lower bit-widths (e.g., 4-bit or 2-bit) offer greater efficiency. Achieving competitive performance at such extremes requires deeper theoretical understanding, particularly of convergence, generalization, quantization noise, and gradient-spike control~\cite{wang2025adagc}. Strengthening these foundations will be key to making ultra low-precision training practical and scalable.

\textbf{Fine-grained scaling strategies.}
Lower precision increases the prevalence of outliers in weight and activation distributions, harming stability and accuracy. Per-channel or per-token scaling helps but is often insufficient. More adaptive or learnable scaling methods can improve robustness but add memory overhead. A careful balance between granularity and efficiency is needed.

\textbf{Optimizer state compression.}
Optimizer states, often kept in high precision, are a major memory bottleneck, especially in large models, where they can exceed parameter size. Compressing or quantizing these states, particularly sensitive components like second-order statistics, is crucial. Research into compression techniques and optimizers inherently suited for low precision is a promising direction.

\textbf{Unified training frameworks for low-precision training.}
There is a lack of modular, scalable frameworks for full-stack low-precision training. This limits reproducibility and adoption. Future toolkits should be hardware-aware, support dynamic precision scheduling, flexible quantization, and enable deployment simulation.

\textbf{Standardized Benchmarks and Evaluation Protocols.}
The field lacks consistent benchmarks, making cross-method comparison difficult. Standardized suites covering various models, datasets, and precision levels are needed to enable fair, reproducible evaluations and better understand trade-offs.

\textbf{Robustness- and safety-aware evaluation.}
Existing studies mainly report training loss, downstream accuracy, memory, and throughput. As low-precision training is increasingly used in instruction tuning, preference alignment, and agentic workflows, future protocols should additionally measure reasoning retention, alignment retention, refusal consistency, tool-call validity, and multi-step tool-use stability. Without such evaluations, it is difficult to judge which low-precision recipes are sufficiently reliable for real deployments.

\textbf{Extending to broader architectures.}
Low-precision training has mostly focused on LLMs, with other foundation models like VLMs, diffusion models, and speech transformers underexplored. These architectures may have different quantization behaviors, offering new challenges and opportunities.

\textbf{Integration with other efficient training paradigms.}
Quantization is often studied in isolation, missing potential synergies with methods like pruning or low-rank approximation. Combining these can further reduce resource demands. Jointly optimizing such techniques could yield more efficient training without sacrificing performance.

%% file: contents/sec_conclusion.tex
\section{Conclusion}
\label{sec:conclusion}

This survey provides an in-depth summarization of existing low-precision training works for LLMs. We begin with preliminaries about different numerical formats, followed by different low-precision components during training and the benefits of low-precision training. Then, categorized by different numerical formats, we introduce existing works with fixed-point and integer numbers, floating-point numbers, and customized formats. We further revisit the literature from cross-cutting perspectives, highlighting both shared numerical challenges and their implications for model reliability. Following this, we briefly introduce QAT approaches, which share some similarities with low-precision training methods to some extent. Lastly, we provide several potential future research directions.

Overall, low-precision training has emerged as a critical technique for reducing the computational and memory costs of training LLMs while often preserving competitive performance under carefully designed recipes. As both hardware and algorithmic support continue to evolve, we anticipate low-precision training will become an integral part of mainstream LLM development. 
Future research may focus on advancing quantization techniques, improving training stability at ultra low precision, and developing unified frameworks that support diverse model architectures. Additionally, efforts toward standard benchmarks and integration with other efficient training paradigms will be crucial for broader adoption and fair evaluation.
Beyond efficiency, the field also needs a clearer understanding of how low-precision training affects model reliability in safety-critical settings.
We hope this survey serves as a valuable reference for researchers and practitioners working toward scalable and efficient training of next-generation LLMs.

%% file: contents/app_fx_details.tex
\section{Extended Method Inventory for Fixed-Point and Integer Training}
\label{app:fx}

\input{fig/forest_fx}

This section provides an extended inventory of low-precision training methods based on fixed-point and integer quantization. Earlier works often used these two terms interchangeably. For conceptual clarity, this survey distinguishes them according to whether the scaling factor $\Delta$ is fixed or data-dependent.
Early studies predominantly employed fixed-point quantization, where a predefined, constant $\Delta$ is used. For a $B$-bit representation with $k$ fractional bits ($k < B$), the quantized value $q$ of an input $x$ is computed as:
\begin{equation}
\begin{aligned}
q &= \text{clip}\left( \text{round}\left( \frac{x}{\Delta} \right), -2^{B-1}, 2^{B-1}-1 \right), \\
\Delta &= 2^{-k}.
\end{aligned}
\end{equation}
While fixed-point quantization offers advantages in terms of computational speed and memory efficiency compared to floating-point representations, its limited numerical range can result in overflow or underflow, potentially degrading model accuracy.
As modern processors and specialized accelerators are increasingly optimized for integer operations, recent research efforts have shifted toward integer quantization techniques for low-precision training.
Integer quantization typically employs symmetric uniform quantization~\cite{krishnamoorthi2018quantizing}, which enhances hardware efficiency. In this scheme, the scaling factor $\Delta$ is data-dependent. Given an input range $(l, u)$, the quantized value is computed as:
\begin{equation}
\begin{aligned}
q &= \text{round}\left( \frac{\text{clip}(x, -c, c)}{\Delta} \right), \\
\Delta &= \frac{c}{2^{B-1} - 1}, \\
c &= \max(|l|, |u|).
\end{aligned}
\end{equation}
In both quantization formats, the original input is approximated by reconstructing the value $\hat{x}$ as:
\begin{equation}
\hat{x} = q \cdot \Delta.
\end{equation}

\IEEEpubidadjcol

During training, the forward and backward propagation processes largely mirror those in full-precision training, with the main difference being the reduced numerical precision of intermediate values. A notable exception is 1-bit quantized training, where the Straight-Through Estimator (STE) is commonly used during backpropagation. This technique bypasses the non-differentiability of the quantization function by approximating gradients as if the operation had preserved full precision. Although heuristic, this workaround enables effective gradient flow while maintaining significant computational efficiency.

The remainder of this section is organized into fixed-point training, integer-based training, and the extreme case of 1-bit quantized training, as summarized in Figure~\ref{fig:fx}.

\subsection{Fixed-Point Training}

Fixed-point methods established some of the earliest viable low-precision training recipes because they preserve simple arithmetic and map naturally to hardware. Across this line of work, the central question is how to compensate for the narrow shared dynamic range through improved rounding, scaling, and precision scheduling.

\textbf{Ordinary fixed-point.}
One early attempt to use fixed-point formats for deep neural network training was introduced by \mcite{Gupta}{guptaagn15}. Their work examined rounding schemes and found that conventional nearest rounding often discards small gradients by rounding them to zero. To address this, they proposed stochastic rounding, defined as:
\begin{align}
	R(x)= \begin{cases}
		\lfloor x \rfloor   &\text{with probability}~p(x),\\
		\lfloor x \rfloor+\delta   &\text{with probability}~1-p(x),
	\end{cases}
\end{align}
with $p(x) = 1 - \frac{x - \lfloor x \rfloor}{\delta}$. This unbiased rounding ($\mathcal{E}(R(x)) = x$) preserves gradient information by assigning a non-zero chance to small updates. Their 16-bit fixed-point training showed minimal performance drop.
\mcite{Xia}{xia2021simple} improved this for binary classification by using a constant $p(x) = 0.5$ for all $x$ in $[\lfloor x \rfloor, \lfloor x \rfloor + \delta)$, enhancing both gradient retention and hardware efficiency. Their 16-bit models converged faster and achieved higher accuracy than with the original method.

\mcite{Chen}{chenhzx17} addressed a key limitation in binarized neural networks where high-precision parameters are maintained. Their FxpNet system represents both weights and gradients in adaptive fixed-point formats. During forward and backward passes, weights, activations, and gradients are quantized, with layer-wise scaling factors adjusted based on overflow thresholds. Using 12-bit representations, FxpNet matches the accuracy of full-precision models.

\mcite{Sakr}{sakrs19} advanced true fixed-point training by proposing a low-precision SGD algorithm. Their method manages quantization noise, precision trade-offs, and dynamic range limits to maintain training stability. They introduced five criteria, including noise equalization, clipped gradients, and minimal quantization bias, to ensure convergence closely mirrors floating-point baselines. Their framework achieved up to 6$\times$ lower representation cost, 8$\times$ computation reduction, and 4$\times$ savings in communication.

To reduce training memory, \mcite{Chakrabarti}{chakrabartim19} proposed compressing activations during training. Exact activations are used for forward computation, but only compressed versions, quantized via fixed-point, are stored for backpropagation. Using as little as 4-bit precision, their approach retained full-precision accuracy while significantly lowering memory usage.

Building on the idea that averaging mitigates quantization noise, \mcite{Yang}{yangzkbws19} introduced SWALP, which averages low-precision SGD iterates using a modified learning rate schedule. All variables, including gradient accumulators, are quantized to 8-bit fixed-point. Theoretical analysis shows convergence for quadratic objectives.
\mcite{Rajagopal}{rajagopalvvb20} proposed MuPPET, a multilevel optimization framework with dynamic precision switching. It combines several fixed-point schemes and gradually increases precision during training. Experiments show MuPPET maintains full-precision accuracy while boosting computational efficiency.

\mcite{Dai}{dai2024trainable} recently introduced QFX, a framework that learns binary point positions during training to optimize FPGA deployment. It uses multiplier-free quantization to reduce DSP usage. Unlike post-training quantization, QFX integrates fixed-point quantization-aware training (QAT), achieving higher accuracy and hardware efficiency.

\textbf{Dynamic fixed-point.}
Unlike standard fixed-point formats, dynamic fixed-point numbers~\cite{williamson1991dynamically} use multiple adjustable scaling factors, striking a balance between fixed- and floating-point formats. \mcite{Courbariaux}{courbariaux2014training} showed their effectiveness for training DNNs. \mcite{Jo}{joplc18} further improved results using weight clipping and batch size scaling with 16-bit dynamic fixed-point.
\mcite{Das}{0002mmkab0vkghd18} extended this work to modern architectures and large datasets. Using ResNet-50, GoogLeNet-v1, VGG-16, and AlexNet on ImageNet-1K, they showed dynamic fixed-point training can match or exceed FP32 accuracy without changing hyperparameters or training schedules.

Fixed-point studies reveal a clear trade-off in that the format is attractive when hardware simplicity and low storage overhead dominate, yet its coarse shared scaling makes it harder to absorb layer-wise distribution shifts and outliers. This limitation largely explains why later work moved toward finer-grained integer quantization, while fixed-point remains especially appealing in structured or FPGA-oriented settings.

\subsection{Integer Training}

Most existing low-precision training methods primarily employ integer formats, with research efforts focusing on different aspects of the training pipeline. To provide a structured overview, we categorize integer training approaches into three primary types:
\begin{itemize}
  \item \textbf{General integer training}, which applies quantization to weights, activations, and gradients throughout the training process;
  \item \textbf{Optimizer-targeted integer training}, which focuses on quantizing optimizer states such as momentum to reduce memory consumption;
  \item \textbf{Communication-targeted integer training}, which aims to enhance the efficiency of distributed training by quantizing the weights or gradients exchanged between devices.
\end{itemize}
Across these branches, the literature evolves from showing that end-to-end integer training is feasible to identifying which tensors, states, and communications need special treatment under aggressive quantization.

\subsubsection{General integer training}

\textbf{Overall design.}
One of the pioneering works in the field of integer training is DoReFa-Net~\cite{zhou2016dorefa}. In addition to using low-bitwidth weights and activations, it adopts stochastic quantization for parameter gradients during backpropagation. This enables the use of bit-level convolution kernels for both training and inference across CPUs, FPGAs, and GPUs. As an early attempt, DoReFa-Net paves the way for accelerating the training of low-bitwidth neural networks.
Building upon this, WAGE~\cite{wulcs18} establishes a framework that discretizes both training and inference by quantizing weights, activations, gradients, and errors to low-bitwidth integers.
\mcite{Yang}{yangdwyxl20} further quantizes optimizer momentum terms, achieving a Fully Quantized Training (FQT) framework.
In parallel, \mcite{Zhang}{zhang2025accurate} address the issue of sparse outlier activations using dynamic block fallback quantization. This method allocates higher bit-widths to outliers when they are detected within specific quantization blocks. Furthermore, \mcite{Zhao}{zhao2024direct} reduce the bit-width to as low as 3, demonstrating the feasibility of ternary weight training.

\textbf{Quantization methods.}
Some studies improve integer quantized training by redesigning quantization schemes to address specific challenges in low-precision optimization.
\mcite{Miyashita}{miyashita2016convolutional} observe that conventional linear quantization struggles with the non-uniform distributions of weights and activations. To address this, they replace fixed-point representation with base-2 logarithmic quantization, which more effectively captures large dynamic ranges using fewer bits. This method demonstrates superior classification accuracy compared to linear quantization.
HALP~\cite{de2018high} tackles the fundamental trade-off between bit-width reduction and quantization noise. It combines Stochastic Variance Reduced Gradient (SVRG) with a dynamic re-centering and re-scaling technique tailored for low-precision numbers. HALP preserves the convergence rate of full-precision SGD despite the presence of quantization noise.
Building upon gradient quantization research, \mcite{Zhao}{zhaohplzgx21} identify a previously overlooked phenomenon, where layer-wise gradients often exhibit multiple distributions along the channel dimensions. To address this, they introduce gradient vectorized quantization, along with a magnitude-aware clipping strategy that minimizes quantization error weighted by gradient magnitudes. The quantization parameters are optimized for each distribution through theoretical derivations.
The dynamic nature of gradients poses challenges for fixed quantization ranges. In-hindsight~\cite{fournarakisn21} reveals that conventional dynamic quantization incurs substantial memory overhead. Their hardware-friendly alternative employs historical quantization ranges updated via moving averages to quantize current tensors. This method leverages accumulator statistics for online range updates, eliminating the need for real-time range computation while maintaining accuracy.
Additionally, Q-GaLore~\cite{zhang2024q} exploits the resilience of gradient projection matrices to low-bit quantization, enabling INT4 quantization of projections while preserving INT8 weights. Stochastic rounding captures gradient accumulation effects, supporting memory-efficient pretraining. Notably, Q-GaLore enables training a LLaMA-7B model from scratch with only 16~GB of GPU memory.

To handle outliers in activation matrices during low-bit quantization, the Hadamard transform~\cite{sylvester1867lx} offers an effective solution by distributing these outliers across all dimensions, making them easier to process. For example, if one dimension contains a very large value while others remain small, applying the transform produces a more balanced vector in which all values are of similar magnitude. This transformation improves the suitability of the data for efficient quantization.
Building on this foundation, \mcite{Xi}{xilcz23} introduce a Hadamard quantizer to suppress activation outliers during forward propagation and exploit the structural sparsity of activation gradients in backpropagation via bit splitting. In this approach, the gradient of each token is separated into higher and lower 4-bit components to prioritize the computation of larger gradients.
Subsequently, \mcite{Schiemer}{schiemer2023hadamard} implement backward-pass matrix multiplications in the Hadamard domain for continual learning scenarios. This enables training with 4-bit integer inputs while maintaining 8-bit accumulators to enhance numerical stability.
\mcite{Kim}{kim2024hlq} develop Hadamard low-rank quantization to optimize the computational cost of backpropagation. Through sensitivity analysis of gradient computations, they design a pipeline that applies 4-bit Hadamard quantization to activation gradients, while combining Hadamard transforms with low-rank approximation techniques for weight gradients.

\textbf{Fine-grained quantization.}
While traditional approaches often use uniform quantization with a global scaling factor, emerging research shows that applying layer-specific or block-specific scaling strategies better accommodates varying data distributions across network components while maintaining model accuracy.
Early work by \mcite{Zhang}{zhanglzlhzggdzc20} establishes key observations about layer-wise data characteristics, such as significant inter-layer distribution variance, dynamic range shifts during training, and the correlation between variance magnitude and required bit-width. Building on these insights, the authors propose a precision-adaptive quantization method that stabilizes data distributions through layer-wise bit-width adjustment. This approach achieves significant training speedup with negligible accuracy loss and no need for hyperparameter modifications.
NITI~\cite{wangrls22} introduces per-layer block exponentiation for dynamic range adaptation, enabling discrete parameter updates for 8-bit integer storage throughout training. It also employs hardware-efficient pseudo-stochastic rounding using intermediate accumulation bits as random sources. By integrating these techniques with an integer-optimized cross-entropy backpropagation approximation, NITI achieves full integer training with minimal computational overhead.
AMPA~\cite{dingfhddlzx24} proposes layer-wise sensitivity measurements for weights, activations, and gradients, dynamically adjusting bit-widths during training and achieving an average precision lower than INT8.
Other complementary approaches explore alternative adaptation strategies. 
\mcite{Shen}{shenlslgll25} propose a capacity-aware method that directly assigns decreasing bit-widths to layers with large capacity, eliminating the need for exhaustive sensitivity analysis while maintaining stability across trials. Meanwhile, FracTrain~\cite{fuyz0lgwl20} automatically adjusted layer precisions based on input characteristics, and Jetfire~\cite{xicztcz24} demonstrates that per-block quantization could effectively preserve accuracy.

\textbf{Quantization errors.}
Several studies focus on mitigating quantization-induced errors through both theoretical and empirical approaches. Through theoretical analysis, \mcite{Banner}{bannerhhs18} show that most training operations are robust to substantial quantization, with only specific gradient computations requiring higher precision. Specifically, they bifurcate the layer gradients, using 16-bit precision for computing the weight gradient, while keeping 8-bit precision for computing the next-layer gradient.
The work by \mcite{Park}{park2018training} introduces an auxiliary parameter to retain partial accumulations of small gradient values, thus addressing precision shortage issues in parameter updates. They also propose a simple guideline for selecting the appropriate bit-width for the final fully connected layer followed by a softmax nonlinearity.

Viewed from this perspective, later methods mainly differ in which source of update distortion they target and what auxiliary mechanism they add to counter it. The study of \mcite{Zhu}{zhugylwlyy20} identifies four distinct gradient distribution characteristics (sharp and wide, evolutionary, depth-specific, and structure-specific) that contribute to greater quantization error. Their solution combines deviation-counteractive learning rate scaling with a cosine-distance-based gradient clipping method.
\mcite{Guo}{guo2024towards} find that proper channel grouping significantly reduces quantization errors. Based on this insight, they propose ShiftQuant, which mitigates gradient quantization errors through intelligent channel grouping.

\textbf{Dynamic precision.}
Some studies explore dynamic precision adjustment during training to balance computational efficiency with model accuracy.
FracTrain~\cite{fuyz0lgwl20} gradually increases the precision of activations, weights, and gradients, reaching standard low-precision levels only in the final stages of training. Their approach also includes automatic adaptation of layer-wise precision. The findings suggest that using lower precision in the early stages of training, followed by higher precision during final convergence, achieves an optimal trade-off between exploratory learning and final model accuracy.
CPT~\cite{fuglydcl21} extends this idea by introducing a cyclic precision schedule, where precision levels vary periodically throughout training. The optimal bounds for these cycles are determined in the early training phase using a simple precision range test.

\textbf{Normalization layers.}
Batch normalization (BN)~\cite{bn} layers, which are an important component in early CNN architectures, require high-precision computation to avoid issues such as zero variance and large dynamic range. To mitigate this drawback, some works adopt full-precision computation for BN layers~\cite{zhou2016dorefa}, while others remove BN layers entirely from the model~\cite{yangcdygl22}. \mcite{Banner}{bannerhhs18} introduces Range BN, which shows significantly higher tolerance to quantization noise. WAGEUBN~\cite{yangdwyxl20} replaces each BN layer with a constant scaling factor. \mcite{Guo}{guo2024towards} develops L1 normalization layers, which are mathematically equivalent to L2-norm batch normalization but demonstrate stronger linearity. Meanwhile, \mcite{Yang}{yangdyxl21} quantizes L1 normalization layers, enabling them to run on integer-based hardware, which typically lacks support for square root operations.

\textbf{Alternative objectives.}
Advances in quantization techniques extend beyond conventional targets such as weights, activations, and gradients to address more specialized challenges in low-precision training.
A notable gap in this area is the handling of full-precision latent weights during integerized training, as identified by \mcite{Fei}{feidzzlzx25}. These latent weights consume substantial memory to accumulate gradient updates for optimizing discrete parameters. To address this, the authors introduce residual quantization to suppress noise in latent weights, along with a dual quantizer that employs optimal nonuniform codebooks to minimize training perturbations.
Beyond weight representation, the computational pipeline of quantization methods presents another major challenge. As \mcite{Xi}{xicztcz24} observe, most quantized training methods adopt a quantize-compute-dequantize paradigm, which proves particularly inefficient for transformer architectures. This frequent data conversion results in significant memory access overhead and degrades the performance of low-precision computation. To address these issues, the authors propose Jetfire, an INT8 training framework featuring redesigned linear and nonlinear operators to support direct INT8 data flow. In addition, they employ per-block quantization to preserve model accuracy. Experiments show that Jetfire achieves accuracy comparable to FP16 training baselines and outperforms existing INT8 training approaches for transformers.

\textbf{Extended applications.}
Some works extend integer training beyond classification tasks. 
\mcite{Yang}{yangdyxl21} propose a new quantization framework for the training and inference of segmentation networks, where both parameters and operations are constrained to 8-bit integer-based values. \mcite{Wortsman}{wortsmandzmfs23} accelerate the training of vision-language models (VLMs) by using 16-bit precision for weight gradients, while employing 8-bit integers for the forward pass and input gradients. Additionally, it adopted AdamW along with the update clipping technique introduced in AdaFactor~\cite{shazeers18}, which tracks the average ratio of the squared gradients to the second moment estimator and reduces the learning rate when this ratio becomes large.

\textbf{Activation compression.}
While gradient compression techniques are well-studied, a specialized line of research focuses on compressing full-precision stored activations specifically for backward passes to reduce memory usage during training, while preserving forward pass precision to maintain model accuracy. These approaches are typically referred to as activation-compressed training (ACT).
Based on the observation that activation values are concentrated in narrow ranges with sparse outliers, \mcite{Park}{parkyv18} apply reduced precision only to the dense regions, thereby minimizing quantization errors for the majority of data while handling outliers separately in high precision. This method achieves 3-bit quantization of activations.
Subsequent works identify several fundamental limitations in ACT, including unclear convergence behavior and non-specific quantization designs~\cite{chenzywsm021}, and architecture-specific restrictions~\cite{liuzwcchcltgmc22}. The framework ActNN~\cite{chenzywsm021} addressed these issues by introducing adaptive quantization with theoretical guarantees. ActNN adopts per-group quantization with dynamic bit allocation, choosing the numerical precision adaptively for each sample and layer. Group range and zero points remain in BF16, and the method minimizes overall variance within a given bit-budget by allocating more bits to sensitive layers.
Building on these insights, \mcite{Liu}{liuzwcchcltgmc22} propose GACT, which extends ACT to diverse model architectures through runtime compression ratio adaptation. GACT estimates the impact of each tensor on the gradient and adjusts the compression ratio accordingly.
Beyond tensor quantization, \mcite{Novikov}{novikovbgsdo23} tackle memory overheads from nonlinearity operations by approximating the derivative of the activation function as piecewise-constant functions. This approximation allows the replacement of full-precision activation storage with bin indices.
In addition to conventional model architectures, solutions for graph neural networks (GNNs) also emerge. 
\mcite{Liu}{liuzylch22} introduce EXACT, the first framework optimized for GPU support of GNN-specific activation compression.
\mcite{Eliassen}{eliassens24} further improve efficiency through block-wise quantization, demonstrating significant reductions in memory consumption.

\textbf{Theory.}
While many empirical studies exist, researchers also seek to establish convergence guarantees for integer quantized training.
Theoretical analyses reveal an intrinsic robustness in neural network training towards precision reduction. 
\mcite{Banner}{bannerhhs18} demonstrate that most training operations can tolerate substantial quantization, with only a few components requiring higher precision. This finding is further supported by \mcite{Li}{lid0ssg17}. The authors identify a critical distinction between convex and non-convex regimes. While convex problems allow quantized training with accuracy guarantees, non-convex landscapes require high-precision representations for effective greedy search phases.
Regarding convergence bounds for quantized training, earlier work suffers from dimension-dependent bounds, as noted by 
\mcite{Li}{lis19}, where the required bit-widths scaled with model dimensionality $d$. Their contribution establishes dimension-independent bounds and extends the analysis to diverse quantization schemes. For specific integer quantization, \mcite{Zhu}{zhugylwlyy20} provides convergence bounds for INT8 training, while \mcite{Zhang}{zhang0kalz17} achieve a 16$\times$ precision reduction in linear models with rigorous guarantees.
Moving beyond worst-case analyses, \mcite{Chen}{chengymg20} introduce a statistical framework for FQT. A key theoretical contribution of this work proves that FQT gradients are unbiased estimators of QAT gradients, implying that FQT and QAT algorithms exhibit the same convergence behavior as the learning rate approaches zero. The study further addresses practical challenges by proposing two novel gradient quantizers that handle dynamic range variation and optimize dimension-wise signal distribution, enabling 5-bit gradient encoding in ResNet50 without any accuracy degradation.

Overall, general integer training has progressed from uniform low-bit quantization to selective schemes that protect outliers, sensitive layers, and expensive data movement. The main lesson is that practical fully quantized training rarely comes from quantizing everything identically; it comes from matching granularity, scaling, and a few high-precision exceptions to tensor statistics and operator roles.

\subsubsection{Optimizer-targeted integer training}

While quantized training alleviates memory pressure during both forward and backward passes, the optimizer states remain a significant bottleneck. With the rapid advancement of deep learning, stateful optimization methods become the de facto standard for training neural networks. Unlike traditional stateless optimizers such as SGD, modern adaptive optimizers like Adam~\cite{adam} utilize historical gradient statistics to dynamically adjust learning rates for each parameter. Although these methods significantly improve convergence and generalization, they introduce substantial memory overhead by maintaining auxiliary variables such as first-order and second-order statistics.

These statistics incur additional memory costs proportional to the number of trainable parameters, leading to a memory footprint for Adam that is 2–3 times larger than that of SGD. In the context of LLMs, this overhead becomes a critical bottleneck. Consequently, there is growing interest in developing efficient optimization techniques that compress optimizer states.

\textbf{Adam.}
Early work by \mcite{Ramesh}{rameshpggvrcs21} demonstrates the feasibility of stable training using 16-bit Adam moments. Building upon this, \mcite{Dettmers}{dettmerslsz22} pioneers the use of 8-bit statistics through block-wise quantization, in which tensors are divided into smaller, independently quantized blocks. Their approach combines dynamic quantization with a stabilized embedding layer to mitigate gradient variance arising from the non-uniform token distributions in language models.
An empirical study by \mcite{Chitsaz}{chitsazfmc24} systematically evaluates various quantization strategies, revealing that while the first-order moments in Adam tolerate 4-bit quantization, the second-order moments require more careful handling, even at 8-bit precision. These findings inform their recommended quantization recipe for pretraining.
Further advancing the field, \mcite{Modoranu}{modoranusmk0ra24} propose MicroAdam, which compresses gradient information before it is incorporated into optimizer states. By employing a novel variant of error correction~\cite{seidefdly14}, and compressing the error feedback itself, their method preserves convergence guarantees while significantly reducing memory consumption.
Pushing the boundaries of low-bit optimization, \mcite{Li}{licz23} achieve 4-bit quantization of optimizer states through two key innovations. First, they introduce adaptive block sizing to effectively handle moment outliers across parameter tensors. Second, they apply a linear quantizer that excludes problematic zero-point values in the second-order moments. Additionally, they incorporate rank-1 normalization, where the scaling factor of each tensor element is determined by the minimum value of the row or column it resides in, allowing for more accurate outlier approximation.

\textbf{Other optimizers.}
While Adam remains the dominant optimizer in model training, emerging research explores state compression for more advanced optimizers such as Lion~\cite{chenlhrw0dlhll23} and Shampoo~\cite{0001ks18}. \cite{li2023qft} propose QFT, a quantized training framework that employs Lion as the optimizer. Lion tracks only momentum and maintains consistent update magnitudes for each parameter, enabling robust integer-based storage, of all optimizer states. In their framework, both gradients and momentum undergo uniform quantization, while weights use a Dense-and-Sparse Quantizer with a specialized gradient flow mechanism for quantized weight updates.

Second-order optimizers like Shampoo pose greater memory challenges due to their use of preconditioning matrices. Several approaches address this constraint through 4-bit quantization. \mcite{Li}{li2024memory} tackle the quantization of non-diagonal preconditioners using Cholesky decomposition and quantization of Cholesky factors with error compensation. Their innovations include a 4-bit exponential moving average (EMA) error estimator and an efficient matrix packing scheme that stores both quantized factors and error states in a single triangular matrix. Theoretical analysis confirms convergence for both smooth and nonsmooth optimization.
Meanwhile, \mcite{Wang}{wangzlh24} identify the eigenvector matrix as a more quantization-friendly target than the preconditioner itself, thereby avoiding distortion of small singular values during inverse root calculations. Their method enhances precision via Björck orthonormalization and shows that linear square quantization yields better results for second-order optimizer states.

This line of work suggests that optimizer compression is constrained less by average precision than by rare but consequential state outliers, especially in second-order statistics. Recent progress therefore relies on block structure, range adaptation, and error correction rather than naive uniform low-bit storage.

\subsubsection{Communication-targeted integer training}

As models continue to grow in scale and complexity, training them efficiently on large datasets has become increasingly dependent on distributed computation across multiple processors and machines. A widely adopted strategy for this purpose is Distributed Data Parallel (DDP), where each node processes a subset of data and synchronizes model updates across all nodes. However, the scalability of such systems is often hampered by communication bottlenecks, especially when full-precision gradients and activations need to be exchanged frequently between nodes. To address this issue, a range of gradient compression techniques have been proposed, aiming to reduce communication overhead while preserving convergence guarantees and training accuracy.

\textbf{DDP.}
One early effort in this direction explores the use of 8-bit approximations for compressing both gradients and activations~\cite{dettmers20158}, achieving up to a 2$\times$ speedup in data transfer compared to standard 32-bit approaches without compromising predictive performance.
Quantized SGD (QSGD)~\cite{alistarhg0tv17} extends this idea by introducing a family of lossy gradient quantization schemes that balance communication efficiency and convergence guarantees, enabling users to control the trade-off between gradient precision and iteration variance. It provides theoretical support for convergence under both convex and non-convex objectives.

Some works develop error compensation mechanisms. The importance of proper error handling is emphasized by \mcite{Li}{libvlxmy24}, who show that naive bidirectional compression schemes incur significant computational overhead and accuracy degradation. To handle this, \mcite{Tang}{tangylzl19} propose DoubleSqueeze, which implements bidirectional compression between workers and parameter servers while maintaining theoretical convergence guarantees. Their analysis show that error-compensated approaches exhibit better tolerance to compression noise compared to naive quantization. Similarly, \mcite{Xie}{xie2025loco} introduce LoCo with a refined error-feedback mechanism using moving averages of past compression errors, demonstrating improved training stability. In addition, \mcite{Chen}{chenshl21} integrates both gradient quantization and weight quantization within the parameter-server model. To address the bias introduced by quantization, the authors propose an error-feedback technique and theoretically establish convergence to first-order stationary points in stochastic nonconvex settings. Efficient-Adam \cite{chensll23} further enhances communication efficiency by introducing a novel two-way quantization scheme combined with a tailored error-feedback strategy.

Recent advances have introduced more sophisticated quantization strategies. \mcite{Yu}{yuwh19} develop a double quantization framework that compresses both parameters and gradients while maintaining the original convergence rate through careful variance analysis. They further combine quantization with sparsification, establishing relationships between sparsity budgets and convergence. \mcite{Faghri}{faghritma0r20} take an adaptive approach, proposing ALQ and AMQ schemes that dynamically adjust to changing gradient statistics during training, where ALQ leverages estimated gradient distributions for optimization, while AMQ utilizes exponentially spaced levels to achieve similar objectives.

At the system level, Tensor Homomorphic Compression (THC)~\cite{libvlxmy24} introduces a preprocessing approach using randomized hadamard transform to enable direct aggregation of compressed values, eliminating decompression overhead. complementing this, \mcite{Wang}{wang2023zero++} present ZeRO++, which integrates block quantization into the Zero Redundancy Optimizer framework, particularly focusing on optimizing all-gather operations through weight quantization.

\textbf{Non-DDP.}
Recent research has extended communication optimization beyond traditional data-parallel settings to accommodate a broader range of parallel training paradigms.
While gradient compression has been extensively explored in data-parallel training, activation compression in pipeline-parallel scenarios remains relatively under-investigated. Addressing this gap, \mcite{Wang}{wangyrhdcr022} introduce AQ-SGD, a novel approach that compresses activation changes rather than their absolute values. This approach demonstrates theoretical convergence guarantees for non-convex optimization while significantly reducing communication costs during pipeline parallel training.

In the context of Sharded Data Parallelism (ShardedDP), \mcite{Jia}{jiaxlwf0slzlt24} propose SDP4Bit with two key innovations. First, it compresses weight updates by quantizing temporal differences between weights, achieving 4-bit representation. Second, it introduces a two-level gradient quantization framework that uses higher precision (8-bit) for intra-node communication and lower precision (4-bit) for inter-node synchronization. The design is further enhanced with a Hadamard Transform to manage gradient outliers.

For large-scale model training, Fully Sharded Data Parallelism (FSDP) has gained traction due to its memory and compute efficiency. Building on this, \mcite{Markov}{markovvga23} develop QSDP, a quantized variant of FSDP. Building on layer-wise parameter gathering in FSDP, QSDP introduces quantization before all-to-all communications, where gradients are compressed using standard unbiased compressors, while weights employ a novel unbiased estimator. 

Compared with compute-side quantization, communication-oriented methods are limited mainly by compression bias, topology effects, and decompression overhead. The field has therefore moved from simple low-bit message passing to error-feedback, topology-aware quantization, and schemes tailored to modern parallel training stacks such as ZeRO, ShardedDP, and FSDP.

\subsection{Binary Training}

\textbf{Binary neural network.}
As a extreme case of integer training, Binary neural networks (BNNs) have emerged as a promising approach for efficient deep learning, particularly in resource-constrained scenarios. By quantizing weights and activations to $\pm1$, BNNs achieve substantial computational savings by replacing multiply-accumulate operations with simpler additions and subtractions.

The foundation of modern BNN methods is established by BinaryConnect~\cite{courbariauxbd15}. This technique maintains full-precision weights during gradient accumulation while using binary values for both forward and backward propagation. It not only reduces computational complexity but also serves as an effective form of regularization. The binarization process can be implemented either deterministically using the sign function:
\begin{align}
	q = \mathrm{sign}(x) =  \begin{cases}
		+1   &\text{if}~x \geq 0,\\
		-1   &\text{otherwise},
	\end{cases}
\end{align}
or stochastically as follows:
\begin{align}
	q = \begin{cases}
		+1   &\text{with probability}~\sigma(x),\\
		-1   &\text{with probability}~1-\sigma(x),
	\end{cases}
\end{align}
where $\sigma(x)$ denotes the computationally efficient hard sigmoid function~\cite{courbariauxbd15}.

Building on this foundation, \mcite{Rastegari}{rastegariorf16} introduce two notable convolutional architectures: Binary-Weight-Networks and XNOR-Networks. The former achieves a 32$\times$ reduction in memory usage by binarizing convolutional filters. The latter extends this approach by also binarizing layer inputs, enabling up to 58$\times$ acceleration in convolutional operations via efficient binary approximations.

Despite these advancements, training BNNs continued to present memory-related challenges. As highlighted by \mcite{Wang}{wangdmzlcccc23}, traditional training methods require storing high-precision activations for all layers, which restricts their deployment on memory-limited devices. To address this, the authors demonstrate that backpropagation is robust to quantization, enabling a memory-efficient training strategy that stores activations solely in binary format without incurring significant accuracy degradation.

The theoretical understanding of binary training is further deepened by \mcite{Gao}{gao20241} by conducting a convergence analysis for 1-bit FQT. Their study show that Adam outperforms SGD in low-bitwidth settings due to its reduced sensitivity to gradient variance. Based on these insights, they introduce two key techniques: Activation Gradient Pruning (AGP), which reduces variance through selective group quantization, and Sample Channel Joint Quantization (SCQ), a hardware-friendly method for efficient gradient computation.

\textbf{Communication speedup.}
1-bit gradient compression has also emerged as a promising solution for reducing bandwidth overhead in distributed deep learning. Early work by \mcite{Seide}{seidefdly14} demonstrate that 1-bit gradient quantization, combined with error feedback by accumulating quantization residuals into subsequent mini-batches can achieve accuracy nearly equivalent to full-precision SGD while significantly lowering communication costs. This principle is generalized by EF-SGD~\cite{karimireddyrsj19}, which establishes convergence guarantees at the same rate as uncompressed SGD for arbitrary compression operators. However, these approaches are primarily designed for SGD-based optimizers, leaving a gap in applicability to adaptive methods such as Adam or LAMB~\cite{lamb}.

To address this limitation, 1-bit Adam~\cite{tanggarlllzh21} introduces a two-stage strategy: an initial warm-up phase using standard Adam to stabilize gradient variance, followed by error-compensated 1-bit compression of momentum while freezing the variance preconditioner. Similarly, \mcite{Lu}{lulzsh23} propose 0/1 Adam, offering provable convergence guarantees under 1-bit quantization constraints.
Extending these ideas to large-batch training, 1-bit LAMB~\cite{liatrh22} adopts a similar warm-up strategy with LAMB before switching to compressed momentum SGD. Despite these advances, practical challenges remain in distributed environments. Notably, \mcite{Wu}{wuh0qwz022} observe that multi-hop all-reduce architectures suffer from cascading compression errors. To counter this, they propose the Marsit framework, which employs unbiased sign aggregation and global compensation to preserve convergence rates in such settings.
To further eliminate practical bottlenecks such as warm-up requirements and the computational overhead of quantization, Birder~\cite{pengqywwl23} provides a solution that natively integrates 1-bit quantization with adaptive updates. It removes the need for full-precision warm-up and theoretically matches the convergence speed of Adam.

Binary training marks the extreme efficiency frontier, but the literature consistently preserves hidden high-precision paths or compensatory mechanisms to control update variance. This recurring pattern underscores how difficult it remains to push all training components to 1-bit simultaneously without selective protection.

%% file: fig/forest_fx.tex
\tikzstyle{leaf}=[
    align=left,
    inner xsep=10pt,
    inner ysep=3pt,
]

\begin{figure*}[htpb]
    \centering
    \begin{forest}
        forked edges,
        for tree={
            l sep=3mm,
            font=\sffamily\bfseries, 
            text=textcolor, 
            grow=east,
            reversed=true,
            anchor=base west,
            parent anchor=east,
            child anchor=west,
            base=center,
            rectangle,
            draw=black,
            rounded corners,
            align=left,
            text centered,
            minimum width=4em,
            edge+={black, line width=1pt},
            s sep=3pt,
            inner xsep=2pt,
            inner ysep=3pt,
            line width=0.8pt,
            ver/.style={rotate=90, child anchor=north, parent anchor=south, anchor=center},
            myline/.style={line width=0.8pt},
        },
        where level=0{fill=rootcolor,text=white,text width=12em,font=\footnotesize,}{},
        where level=1{fill=level1color,text=white,text width=7em,font=\footnotesize,}{},
        where level=2{fill=level2color,text width=10em,font=\scriptsize,}{},
        where level=3{fill=level3color,text width=26em,font=\scriptsize,leaf,}{},
        [
        {Fixed-Point and Integer Training}, ver,
                [
                {Fixed-Point Training}, name=fx_root
                    [, phantom
                        [
                        {
                        \textbf{Ordinary Fixed-Point:} \mcite{Gupta}{guptaagn15}, 
                        \mcite{Xia}{xia2021simple}, 
                        FxpNet~\cite{chenhzx17}, 
                        \mcite{Sakr}{sakrs19},\\
                        \mcite{Chakrabarti}{chakrabartim19}, 
                        SWALP~\cite{yangzkbws19}, 
                        MuPPET~\cite{rajagopalvvb20}, 
                        QFX~\cite{dai2024trainable}\\
                        \textbf{Dynamic Fixed-Point:} \mcite{Courbariaux}{courbariaux2014training}, 
                        \mcite{Jo}{joplc18}, 
                        \mcite{Das}{0002mmkab0vkghd18}
                        }
                        ,
                        name=fx_leaf
                        ]
                    ]
                ]
                [
                {Integer Training},
                    [
                    {General methods},
                        [
                        {
                        \textbf{Overall Design:} DoReFa-Net~\cite{zhou2016dorefa}, 
                        WAGE~\cite{wulcs18}, 
                        WAGEUBN~\cite{yangdwyxl20},
                        \mcite{Zhang}{zhang2025accurate},\\
                        DQT~\cite{zhao2024direct}\\
                        \textbf{Quantization Methods:} \mcite{Miyashita}{miyashita2016convolutional},
                        HALP~\cite{de2018high},
                        \mcite{Zhao}{zhaohplzgx21},\\
                        In-hindsight~\cite{fournarakisn21},
                        Q-GaLore~\cite{zhang2024q},
                        \mcite{Xi}{xilcz23},
                        HDQT~\cite{schiemer2023hadamard},
                        HLQ~\cite{kim2024hlq}\\
                        \textbf{Fine-grained Quantization:} \mcite{Zhang}{zhanglzlhzggdzc20},
                        NITI~\cite{wangrls22},
                        AMPA~\cite{dingfhddlzx24},\\
                        \mcite{Shen}{shenlslgll25},
                        FracTrain~\cite{fuyz0lgwl20},
                        Jetfire~\cite{xicztcz24}\\
                        \textbf{Quantization Errors:} \mcite{Banner}{bannerhhs18},
                        \mcite{Park}{park2018training},
                        \mcite{Zhu}{zhugylwlyy20},
                        \mcite{Guo}{guo2024towards}\\
                        \textbf{Dynamic Precision:} FracTrain~\cite{fuyz0lgwl20},
                        CPT~\cite{fuglydcl21}\\
                        \textbf{Normalization Layers:} \mcite{Banner}{bannerhhs18},
                        WAGEUBN~\cite{yangdwyxl20},
                        \mcite{Yang}{yangdyxl21}\\
                        \textbf{Alternative Objectives:} \mcite{Fei}{feidzzlzx25},
                        Jetfire~\cite{xicztcz24}\\
                        \textbf{Extended Applications:} \mcite{Yang}{yangdyxl21},
                        SwitchBack~\cite{wortsmandzmfs23}\\
                        \textbf{Activation Compression:} \mcite{Park}{parkyv18},
                        ActNN~\cite{chenzywsm021},
                        GACT~\cite{liuzwcchcltgmc22},
                        \mcite{Novikov}{novikovbgsdo23},\\
                        EXACT~\cite{liuzylch22},
                        \mcite{Eliassen}{eliassens24}\\
                        \textbf{Theory:} \mcite{Banner}{bannerhhs18},
                        \mcite{Li}{lid0ssg17},
                        \mcite{Li}{lis19},
                        \mcite{Zhu}{zhugylwlyy20},
                        ZipML~\cite{zhang0kalz17},\\
                        \mcite{Chen}{chengymg20}
                        }
                        ,
                        ]
                    ]
                    [
                    {\makecell{Optimizer-targeted methods}},
                        [
                        {
                        \textbf{Adam:} DALL-E~\cite{rameshpggvrcs21},
                        \mcite{Dettmers}{dettmerslsz22},
                        \mcite{Chitsaz}{chitsazfmc24},
                        MicroAdam~\cite{modoranusmk0ra24},\\
                        \mcite{Li}{licz23}\\
                        \textbf{Other Optimizers:} QFT~\cite{li2023qft},
                        \mcite{Li}{li2024memory},
                        \mcite{Wang}{wangzlh24}
                        }
                        ,
                        ]
                    ]
                    [
                    {Communication-targeted methods},
                        [
                        {
                        \textbf{DDP:} \mcite{Dettmers}{dettmers20158},
                        QSGD~\cite{alistarhg0tv17},
                        DoubleSqueeze~\cite{tangylzl19},
                        LoCo~\cite{xie2025loco},
                        \mcite{Chen}{chenshl21},\\
                        Efficient-Adam \cite{chensll23},
                        AsyLPG~\cite{yuwh19},
                        \mcite{Faghri}{faghritma0r20},
                        THC~\cite{libvlxmy24},
                        ZeRO++~\cite{wang2023zero++}\\
                        \textbf{Non-DDP:} AQ-SGD~\cite{wangyrhdcr022},
                        SDP4Bit~\cite{jiaxlwf0slzlt24},
                        QSDP~\cite{markovvga23}
                        }
                        ,
                        ]
                    ]
                ]
                [
                {Binary Training}, name=bn_root
                    [, phantom
                        [
                        {
                        \textbf{Binary Neural Networks:} BinaryConnect~\cite{courbariauxbd15},
                        XNOR-Net~\cite{rastegariorf16},
                        \mcite{Wang}{wangdmzlcccc23},\\
                        1-Bit FQT~\cite{gao20241}\\
                        \textbf{Communication Speedup:} \mcite{Seide}{seidefdly14},
                        EF-SGD~\cite{karimireddyrsj19},
                        1-bit Adam~\cite{tanggarlllzh21},\\
                        0/1 Adam~\cite{lulzsh23},
                        1-bit LAMB~\cite{liatrh22},
                        \mcite{Wu}{wuh0qwz022},
                        Birder~\cite{pengqywwl23}
                        }
                        ,
                        name=bn_leaf
                        ]
                    ]
                ]
        ]
        \draw[line width=1pt] (fx_root) -- (fx_leaf);
        \draw[line width=1pt] (bn_root) -- (bn_leaf);
    \end{forest}
    \caption{Overview of studies on training with low-precision fixed-point and integer formats.}
    \label{fig:fx}
\end{figure*}

%% file: contents/app_fp_details.tex
\section{Extended Method Inventory for Floating-Point Training}
\label{app:fp}

\input{fig/forest_fp}

This section provides an extended inventory of low-precision training methods based on floating-point quantization. Converting a high-precision floating-point number to a lower-precision one follows a different numerical process from integer quantization. Specifically, for converting a high-precision floating-point value (E$e$M$m$) to a lower-precision representation (E${e'}$M${m'}$), we begin by copying the lower $e'$ exponent bits from the source to the target. The mantissa is then truncated to $m'$ bits by rounding to the nearest value. To better preserve information during quantization, a scaling factor $\Delta$ is typically applied to the source value prior to conversion.
If an overflow occurs, the result is clipped directly to the maximum or minimum representable value. In the case of underflow, the value is divided by the smallest subnormal number in the low-precision format, rounded to the nearest integer, and then multiplied back by the same smallest subnormal number.
Low-precision floating-point training does not require specialized backward propagation rules in the same way as binarized training. Instead, it largely follows the full-precision training process, with the main difference being the reduced numerical precision.

This section first reviews widely used 16-bit floating-point training techniques and then turns to lower-precision approaches based on 8-bit and 4-bit floating-point formats. The overall structure is illustrated in Figure~\ref{fig:fp}.

Compared with integer training, floating-point methods are driven more directly by range management. As precision drops from 16-bit to 8-bit and 4-bit, progress depends less on the nominal format alone than on how scaling, accumulation, and selective high-precision exceptions are coordinated.

\subsection{16-Bit Floating-Point Training}

The adoption of 16-bit floating-point formats for deep learning training has gained widespread popularity, driven by their computational efficiency and memory savings. Modern deep learning frameworks support these formats extensively, with FP16 and BF16 emerging as the most prominent choices.

Early work by \mcite{Micikevicius}{micikeviciusnad18} pioneered FP16 training by addressing its limited dynamic range with three key techniques: maintaining an FP32 master copy of weights for accurate updates, applying loss scaling to prevent gradient underflow, and using FP32 accumulation for partial products. Together, these strategies enabled FP16 to match FP32 accuracy across various architectures.

However, \mcite{Zhao}{zhaoval20} observed that uniform loss scaling struggles with varying gradient distributions across layers, often requiring heuristic tuning. To address this, they proposed a dynamic gradient scaling method that adjusts per-tensor scales during backpropagation, keeping the underflow rate below a set threshold and avoiding overflow, thereby improving stability.

Precision conversion performance has also been studied. \mcite{He}{he00022} found that format casting overhead can offset benefits of FP16, taking over 21\% of execution time in some cases. To reduce this, they introduced Campo, a cost-aware graph rewriting framework that applies FP16 only where it offers net speedup. By modeling performance, Campo cuts casting overhead and boosts efficiency.

BF16 has also emerged as a stable alternative. As shown by \mcite{Kalamkar}{kalamkar2019study}, it matches dynamic range of FP32, removing the need for hyperparameter tuning and simplifying precision conversion. Results confirm BF16 can match convergence of FP32 across domains.

As a result, BF16 has gradually replaced FP16 as the default format for 16-bit training. Nevertheless, with the advancement of modern GPUs, researchers are now exploring even more efficient training methods that leverage floating-point formats with lower precisions.

\subsection{Sub-8-Bit Floating-Point Training}

\textbf{Overall design.}
The adoption of lower numerical precision for model training presents significant challenges due to the dramatically reduced representation range. Early attempts at FP8 training show substantial accuracy degradation in popular architectures such as MobileNet and Transformers, primarily due to differing precision requirements between the forward and backward passes \cite{sunccwvsczg19}. To address this, \mcite{Sun}{sunccwvsczg19} propose a hybrid format approach, using E4M3 for the forward pass and E5M2 for the backward pass. This method enables successful training across various tasks without accuracy loss. \mcite{Noune}{noune20228} conduct a systematic empirical study of 8-bit formats, identifying optimal exponent and mantissa configurations that preserve accuracy while enhancing training speed and power efficiency across multiple domains.
\mcite{Lee}{lee2024fp8} investigate the robustness of reduced-precision training for LLMs, highlighting the instability of current FP8 methods and proposing new evaluation techniques and a sharpness-based metric to assess training stability under varying precision levels. Their analysis aims to guide the development of more reliable and cost-effective low-precision training schemes.
In terms of optimizing communication in distributed training, \mcite{Han}{han2021auto} propose Auto Precision Scaling (APS), a method that enables accurate and efficient distributed deep learning by communicating gradients in 8-bit floating-point values, achieving minimal or no accuracy loss and significant speedups. 

In the field of hardware design, \mcite{Desrentes}{desrentes2023exact} propose architectures for exact dot product accumulate operators tailored to 8-bit floating-point formats, enabling precise accumulation by expanding products to fixed-point and rounding from wide accumulators. 
\mcite{Lutz}{lutz2024fused} present two novel microarchitectures for fused FP8 DOT4 operations with single rounding, targeting efficient GEMM in ML workloads by accumulating to FP32 with dynamic range scaling. Their designs, late accumulation and early accumulation, optimize power and area efficiency.
Additionally, HiFloat8 Format for Ascend architectures has also been studied by \mcite{Luo}{luo2024ascend}.

Research in low-precision floating-point training progressively advances toward more aggressive quantization strategies. \mcite{Sun}{sunwcnacvmsg20} pioneer 4-bit training using Radix-4 FP4 formats (E3M0) combined with specialized rounding schemes. These schemes use nearest rounding and select the midpoint between two neighboring exponent levels as the rounding threshold. Gradient scaling techniques are also employed. Since each output activation gradient is used twice during backpropagation, the authors quantize it to either the even or odd phase to compute both the input activation gradient and the weight activation gradient, mitigating expected quantization error through cancellation. Furthermore, they analyze the quantization bias effects on batch normalization, finding that aggressive quantization induces internal covariate shifts, leading to generalization issues.
\mcite{Wang}{wang2025optimizing} extend this direction by introducing differentiable quantization estimators for weights. They derive a function with correction terms for accurate gradient estimation together with outlier compensation strategies for activations, including a clamping method and a sparse auxiliary matrix, which help preserve quantization accuracy and maintain model performance. Their framework specifically targets LLMs.

Recognizing the varying sensitivity across network components in transformer-based architectures, \mcite{Zhou}{zhou2025towards} develop mixed-precision techniques. They adopt FP4 for most operations while retaining FP8 precision for QKV computations and output projections. \mcite{Fishman}{fishman2024scaling} identify prolonged training instabilities in FP8 implementations and attribute these to activation function behaviors. They propose Smooth-SwiGLU, which enables stable training of models with up to 7B parameters.

At scale, \mcite{Micikevicius}{micikevicius2022fp8} empirically validate that FP8 training achieves accuracy comparable to FP16 and bfloat16 across model sizes up to 175B parameters, without requiring hyperparameter adjustments. The practical deployment of these methods is demonstrated by FP8-LM~\cite{peng2023fp8}, which incorporates automatic tensor-wise scaling and supports distributed parallel training. This results in a 39\% memory reduction and a 75\% training speedup compared to BF16 when training a model with 175B parameters.

Recent advances in low-precision training have achieved significant breakthroughs with the introduction of DeepSeek-V3~\cite{liu2024deepseek}, which represents the first successful application of FP8 training in industrial-scale scenarios. This innovative approach incorporates several key technical contributions that collectively enable stable and efficient FP8 training.
At the core of this methodology lies a fine-grained quantization techniques. Specifically, activations are processed in 1$\times$128 tiles and weights in 128$\times$128 blocks, with online calculation of maximum absolute values for each block. 
Due to the fine-grained quantization scheme general matrix multiplication (GEMM) operations are executed entirely in FP8 precision using the E4M3 format for both forward and backward passes. 
Memory efficiency is further enhanced through careful management of cached activations stored in FP8 for backward computation, with specialized designs for specific components. Specifically, the Linear operator inputs following attention use customized E5M6 format, while SwiGLU operator inputs in MoE layers are cached in FP8 and recomputed during backward passes.
To optimize communication in MoE architectures, the system quantizes activations before MoE up-projections and activation gradients before down-projections into FP8. Notably, certain critical components retain higher precision, including the embedding module, output head, MoE gating modules, normalization operators, and attention operators. The framework maintains master weights in FP32, weight gradients in FP32, and optimizer states in BF16, ensuring numerical stability while benefiting from reduced precision where applicable.

\textbf{Value scaling.}
The DeepSeek-V3 example above reflects a broader pattern in low-precision floating-point training: once precision drops to FP8 or FP4, successful recipes depend critically on how values are scaled before quantization. Earlier studies therefore devoted substantial effort to designing scaling strategies that balance dynamic range against effective resolution.

Against this backdrop, existing methods differ mainly in how broadly the scale is shared, how dynamically it is updated, and whether outliers are handled implicitly or explicitly. Early approaches primarily focus on loss scaling techniques. \mcite{Mellempudi}{mellempudi2019mixed} proposes an adaptive method that dynamically adjusts the scaling factor update frequency by monitoring loss progression, effectively compensating for the reduced subnormal range in 8-bit floating-point formats. Building on this, GradScale~\cite{sunwcnacvmsg20} introduces a trainable per-layer gradient scaling, which learns optimal scaling during training.

\mcite{Cambier}{cambierbgent20} presents S2FP8, an innovative 8-bit format that employs shifted and squeezed factors to rescale tensor ranges before truncation, thereby eliminating the need for manual loss scaling tuning. Following this, \mcite{Chmiel}{chmielbshbs21} explores per-layer gradient scaling and identifies optimal values that enable successful quantization of gradients using 6-bit floating-point formats.
\mcite{Blake}{blake2023unit} introduce unit scaling, a method that enables stable low-precision training by ensuring unit variance across weights, activations, and gradients at initialization, eliminating the need for scale tuning or added computational cost.

Recent research shifts toward automation and system-level optimization. FP8-LM~\cite{peng2023fp8} proposes a global scaling strategy that coordinates tensor-wise scaling factors across GPUs using a single shared scalar, streamlining distributed training. \mcite{Perez}{perez2023training} develops a dynamic per-tensor scaling methodology for linear layers, validated on LLMs with up to 70 billion parameters.  Scalify~\cite{balancca2024scalify} propagates scaling information throughout computational graphs. This framework unifies FP8 and FP16 techniques under an automated paradigm through specialized operations and propagation rules.

\textbf{Result accumulation.}
While low-precision representations for weights and activations show promise, maintaining gradient fidelity during backpropagation remains a key challenge, particularly for accumulations in partial product computations and weight updates.
\mcite{Mellempudi}{mellempudi2019mixed} reduces the precision of the master weight copy from 32-bit to 16-bit without compromising model performance. \mcite{Wang}{wangcbcg18} demonstrates successful end-to-end 8-bit floating-point training through two key innovations: chunk-based accumulation, which decomposes long dot products into intra-chunk partial sums followed by inter-chunk aggregation, and floating-point stochastic rounding, which helps preserve gradient fidelity.
Subsequent research shifts toward establishing theoretical foundations for precision requirements. \mcite{Sakr}{sakrwccasg19} develops a statistical model that correlates accumulation length with minimum bit-width by analyzing variance preservation in ensembles of partial sums. This provides principled guidance for allocating precision in low-bit training.
More recently, \mcite{Ali}{alifs24} designs a dedicated FP8 multiply-accumulate (MAC) unit that combines stochastic rounding with optimized 12-bit accumulations, striking a balance between computational efficiency and numerical accuracy.

\textbf{Optimizer states.}
While recent frameworks accelerate training using low-precision floating-point numbers, they often overlook optimizer state compression, leaving substantial memory savings unrealized. To address this gap, FP8-LM~\cite{peng2023fp8} proposes an FP8 mixed-precision optimization scheme that reduces memory consumption from 16 bytes to 6 bytes per parameter. This is achieved by allocating 2 bytes for master weights, 1 byte each for gradients and first-order states, and 2 bytes for second-order states. \mcite{Fishman}{fishman2024scaling} specifically targets moment tensors, introducing dedicated FP8 formats (E4M3 for first moment, E5M2 for second moment) to ensure numerical stability during quantization.

However, naive FP8 quantization of optimizer states leads to under-utilization of the representation range, resulting in suboptimal compression. COAT~\cite{xi2024coat} addresses this limitation through two key innovations. First, it applies dynamic range expansion to optimizer states, using an expand function to align state distributions with the dynamic range of the E4M3 format prior to quantization. Second, it employs mixed-granularity activation quantization, combining per-tensor and per-group strategies to achieve additional memory savings. This co-design of optimizer and activation compression enables significantly greater memory reduction.

\textbf{Block floating-point.}
Based on the standard floating-point format, block floating point (BFP) emerges as an alternative that shares exponents across tensor blocks. This design preserves a wide dynamic range while enabling efficient fixed-point logic for multiply-and-accumulate operations.

Building on this concept, \mcite{Drumond}{drumondljf18} proposes Hybrid BFP (HBFP), an approach that uses BFP for dot products while retaining floating-point arithmetic for other operations. Extending this work, Accuracy Booster~\cite{harma2022accuracy} investigates the precision requirements of HBFP and demonstrates that 6-bit mantissas are sufficient to achieve FP32-level accuracy when applied consistently across all layers and training epochs.

Overall, sub-8-bit floating-point training has evolved from format exploration to full-stack co-design. The frontier is no longer whether FP8 or FP4 arithmetic can be used at all, but how much of the pipeline can be pushed down while retaining fine-grained scaling, protected states, and efficient kernels.

%% file: fig/forest_fp.tex
\tikzstyle{leaf}=[
    align=left,
    inner xsep=10pt,
    inner ysep=3pt,
]

\begin{figure*}[htpb]
    \centering
    \begin{forest}
        forked edges,
        for tree={
            font=\sffamily\bfseries, 
            text=textcolor, 
            grow=east,
            reversed=true,
            anchor=base west,
            parent anchor=east,
            child anchor=west,
            base=center,
            rectangle,
            draw=black,
            rounded corners,
            align=left,
            text centered,
            minimum width=4em,
            edge+={black, line width=1pt},
            s sep=3pt,
            inner xsep=2pt,
            inner ysep=3pt,
            line width=0.8pt,
            ver/.style={rotate=90, child anchor=north, parent anchor=south, anchor=center},
            myline/.style={line width=0.8pt},
        },
        where level=0{fill=rootcolor,text=white,text width=9em,font=\footnotesize,}{},
        where level=1{fill=level1color,text=white,text width=12em,font=\footnotesize,}{},
        where level=2{fill=level3color,text width=26em,font=\scriptsize,leaf,}{},
        [
        {Floating-Point Training}, ver,
                [
                {16-Bit Floating-Point Training}, name=fp16_root
                    [
                    {
                    \textbf{FP16:} \mcite{Micikevicius}{micikeviciusnad18},
                    \mcite{Zhao}{zhaoval20},
                    Campo~\cite{he00022}\\
                    \textbf{BF16:} \mcite{Kalamkar}{kalamkar2019study}
                    }
                    ,
                    name=fp16_leaf,
                    edge={draw=none}
                    ]
                ]
                [
                {Sub-8-Bit Floating-Point Training}, name=sub8_root
                    [
                    {
                    \textbf{Overall Design:} HFP8~\cite{sunccwvsczg19},
                    \mcite{Noune}{noune20228},
                    \mcite{Lee}{lee2024fp8},
                    APS~\cite{{han2021auto}},\\
                    \mcite{Desrentes}{desrentes2023exact},
                    \mcite{Lutz}{lutz2024fused},
                    \mcite{Luo}{luo2024ascend},
                    Radix-4 FP4~\cite{sunwcnacvmsg20},\\
                    \mcite{Wang}{wang2025optimizing},
                    \mcite{Zhou}{zhou2025towards},
                    \mcite{Fishman}{fishman2024scaling},
                    \mcite{Micikevicius}{micikevicius2022fp8},\\
                    FP8-LM~\cite{peng2023fp8},
                    DeepSeek-V3~\cite{liu2024deepseek}\\
                    \textbf{Value Scaling:} \mcite{Mellempudi}{mellempudi2019mixed},
                    GradScale~\cite{sunwcnacvmsg20},
                    S2FP8~\cite{cambierbgent20},
                    \mcite{Chmiel}{chmielbshbs21},\\
                    \mcite{Blake}{blake2023unit},
                    FP8-LM~\cite{peng2023fp8},
                    \mcite{Perez}{perez2023training},
                    Scalify~\cite{balancca2024scalify}\\
                    \textbf{Result Accumulation:} \mcite{Mellempudi}{mellempudi2019mixed},
                    \mcite{Wang}{wangcbcg18},
                    \mcite{Sakr}{sakrwccasg19},\\
                    \mcite{Ali}{alifs24}\\
                    \textbf{Optimizer States:} FP8-LM~\cite{peng2023fp8},
                    \mcite{Fishman}{fishman2024scaling},
                    COAT~\cite{xi2024coat}\\
                    \textbf{Block Floating-Point:} HBFP~\cite{drumondljf18},
                    Accuracy Booster~\cite{harma2022accuracy}
                    }
                    ,
                    name=sub8_leaf,
                    edge={draw=none}
                    ]
                ]
        ]
        \draw[line width=1pt] (fp16_root) -- (fp16_leaf);
        \draw[line width=1pt] (sub8_root) -- (sub8_leaf);
    \end{forest}
    \caption{Overview of studies on training with low-precision floating-point formats.}
    \label{fig:fp}
\end{figure*}

%% file: contents/app_custom_details.tex
\section{Extended Method Inventory for Custom Numerical Formats}
\label{app:custom}

\input{fig/forest_custom}

This section covers customized numerical formats proposed for low-precision training. Certain customized formats, such as NormalFloat~\cite{dettmersphz23}, are designed for pretrained fixed parameters that participate only in the forward stage during training, so they are excluded here because they do not directly address low-precision training. Figure~\ref{fig:custom} presents the structure of this section. Although the encodings differ, most of these formats explore the same design space by determining how much exponent information should be shared, where scaling should be attached, and how hardware simplicity should be traded against adaptability to nonuniform tensor statistics.

\textbf{Posit.}
Posit~\cite{gustafsony17} is an alternative to traditional floating-point formats, offering fixed-bit hardware efficiency along with benefits such as greater dynamic range, improved accuracy, and consistent bitwise results across platforms. Its dynamic segmentation, comprising sign, regime, exponent, and fraction bits, encodes only what's needed, avoiding typical overflow/underflow issues.
\mcite{Dinechin}{de2019posits} evaluate posit vs. floating-point trade-offs, suggesting posit as an effective storage format that blends the strengths of both. \mcite{Lu}{lufxlw21} show that 8-bit posit, combined with tensor-wise scaling, achieves training accuracy comparable to higher-precision floating-point formats.

\textbf{Flexpoint.} 
Flexpoint~\cite{kosterwwnbcehhk17} fuses fixed- and floating-point advantages by associating each tensor with a shared, dynamically adjusted exponent. This approach maximizes dynamic range and prevents overflow by tracking historical max values.
Flexpoint is especially hardware-friendly: it reduces memory and bandwidth usage by amortizing exponent representation and communication across entire tensors.

\textbf{FloatSD.} 
The floating-point signed digit (FloatSD) format~\cite{linskc19} targets CNN weight compression by reducing the number of non-zero digits, simplifying convolutions to additions of shifted multiplicands.
It also applies quantization to mantissa and exponent fields of activations and gradients, using 8-bit floating-point values, which reduces computational complexity during training.

\textbf{MLS.}
The Multi-Level Scaling (MLS) format~\cite{zhongndzzzwy22} aims to balance representational power and energy efficiency. MLS uses element-wise scaling for enhanced dynamic range and a shared group scaling factor to reduce bitwidth, enabling integer-only accumulation with low overhead.
Integrated into a low-bit tensor convolution unit, MLS improves both energy and computational efficiency during training.

\textbf{Microscaling.}
The Microscaling (MX) format~\cite{rouhani2023microscaling} uses per-block scaling factors with low-precision scalar values. Each MX block contains $k$ elements sharing a scaling factor (typically E8M0) and reduced-precision types (FP8, FP6, FP4, or INT8), leading to variants like MXFP8 and MXINT8.
\mcite{Rouhani}{rouhani2023microscaling} show that even 4-bit MX formats can train large transformers with minor accuracy loss.
Expanding on this, \mcite{Tseng}{tseng2025training} present near-lossless training with MXFP4 GEMMs, achieving 2$\times$ speedup over FP8. Their method combines stochastic rounding with a Hadamard transform to reduce block-level variance.
\mcite{Chen}{chen2025oscillation} identify weight oscillation during the forward pass as a key source of MXFP4 degradation. They propose two fixes: Q-EMA, an exponential moving average quantizer that stabilizes rounding, and Q-Ramping, an optimizer that adaptively tunes update frequency for oscillating weights.
More recently, \mcite{Hu}{hu2025elucidating} provide a systematic design-space study of FP4 training under microscaling quantization. Their analysis shows that carefully combining Hadamard transforms, tensor scaling, and stochastic rounding yields the most favorable performance-overhead trade-off, and further suggests UE5M3 as a promising choice for scaling factors.

Overall, custom formats are best viewed as targeted operating points between fixed-point and floating-point extremes rather than a disconnected set of tricks. Their strongest results appear when the encoding is paired with matching scaling rules and hardware or dataflow assumptions, as illustrated most clearly by recent microscaling-based FP4 training.

%% file: fig/forest_custom.tex
\tikzstyle{leaf}=[
    align=left,
    inner xsep=10pt,
    inner ysep=3pt,
]

\begin{figure*}[htpb]
    \centering
    \begin{forest}
        forked edges,
        for tree={
            font=\sffamily\bfseries, 
            text=textcolor, 
            grow=east,
            reversed=true,
            anchor=base west,
            parent anchor=east,
            child anchor=west,
            base=center,
            rectangle,
            draw=black,
            rounded corners,
            align=left,
            text centered,
            minimum width=4em,
            edge+={black, line width=1pt},
            s sep=3pt,
            inner xsep=2pt,
            inner ysep=3pt,
            line width=0.8pt,
            ver/.style={rotate=90, child anchor=north, parent anchor=south, anchor=center},
            myline/.style={line width=0.8pt},
        },
        where level=0{fill=rootcolor,text=white,text width=10em,font=\footnotesize,}{},
        where level=1{fill=level3color,text width=32em,font=\scriptsize,leaf,}{},
        [
        {Custom Numerical Formats}, ver,
                [
                    {
                    \textbf{Posit:} \mcite{Gustafson}{gustafsony17},
                    \mcite{Dinechin}{de2019posits},
                    \mcite{Lu}{lufxlw21}\\
                    A posit number consists of a sign bit, dynamically sized regime bits, and optional exponent and fraction bits.
                    }
                    ,
                ]
                [
                    {
                    \textbf{Flexpoint:} \mcite{K{\"{o}}ster}{kosterwwnbcehhk17}\\
                    Flexpoint uses tensors with a dynamically adjusted shared exponent, balancing precision and dynamic range.
                    }
                    ,
                ]
                [
                    {
                    \textbf{FloatSD:} \mcite{Lin}{linskc19}\\
                    FloatSD reduces convolution operations to the addition of two shifted partial products by limiting non-zero\\
                    digits to two during training.
                    }
                    ,
                ]
                [
                    {
                    \textbf{MLS:} \mcite{Zhong}{zhongndzzzwy22}\\
                    MLS uses element-wise and group scaling to optimize dynamic range and simplify integer accumulation.
                    }
                    ,
                ]
                [
                    {
                    \textbf{Microscaling:} \mcite{Rouhani}{rouhani2023microscaling},
                    \mcite{Tseng}{tseng2025training},
                    \mcite{Chen}{chen2025oscillation},
                    \mcite{Hu}{hu2025elucidating}\\
                    The MX format combines per-block scaling factors (typically in E8M0) with low-precision element-wise\\
                    values (e.g., FP8, INT8) to create efficient variants like MXFP8 or MXINT8.
                    }
                    ,
                ]
        ]
    \end{forest}
    \caption{Overview of studies employing custom numerical formats for low-precision training.}
    \label{fig:custom}
\end{figure*}

%% file: contents/app_qat_details.tex
\section{Extended Method Inventory for Quantization-Aware Training}
\label{app:qat}

This section briefly reviews quantization-aware training (QAT) methods that are closely related to low-precision training. While low-precision training reduces the precision of both the forward and backward passes during optimization, low-precision inference reduces the precision of weights and activations only in the forward pass through quantization.

Methods for low-precision inference are commonly divided into Post-Training Quantization (PTQ) and QAT. PTQ quantizes a pre-trained model without further optimization, usually through calibration heuristics or explicit reconstruction objectives. QAT instead incorporates quantization into training, often via fake-quantization operators or learned quantization parameters, allowing the model to adapt to deployment-time low-precision arithmetic during optimization~\cite{jacob2018quantization,krishnamoorthi2018quantizing,esser2020learned}.Although there is a large body of work on PTQ techniques for LLMs~\cite{wei2022outlier,xiao2023smoothquant,yang2025raana,lin2024awq,zhao2024atom,liu2025septq,zhuz00l0t23,li2023fp8}, those methods are outside the scope of this survey. The remainder of this section therefore focuses on representative QAT studies, organized as shown in Figure~\ref{fig:qat}.

\input{fig/forest_qat}

\textbf{Binary and ternary QAT.}
QAT for LLMs has seen progressive advancements in extreme quantization, starting with the pioneering BitNet architecture~\cite{wang2023bitnet}. As the first 1-bit Transformer for LLMs, BitNet employs binary weights and 8-bit activations while maintaining high-precision gradients during training. Its simple implementation, which replaces linear projections with signum-binarized weights and uses STE-based backpropagation, demonstrates the feasibility of extreme quantization. 
BitNet b1.58~\cite{ma2024era} further introduce ternary parameters through absmean quantization, as follows:
\begin{equation}
    W_q = \max\left(-1, \min\left(1, \mathrm{round}\left(\frac{W}{\gamma + \epsilon}\right)\right)\right),
\end{equation}
where $\gamma=\frac{1}{nm}\sum_{ij} |W_{ij}|$ indicates absolute value of all elements in $w$. 
It retains the activation scheme from the original BitNet. \mcite{Nielsen}{nielsen2025continual} enhances this framework with a progressive two-phase 16-to-1.58-bit training strategy. This approach combines high-precision pre-training followed by a transition to lower-bit quantization, providing a smoother shift for the model and helping it retain more of the knowledge acquired during full-precision training.

\textbf{QAT with KD.}
Several studies incorporate knowledge distillation~\cite{kd} into the QAT process. LLM-QAT~\cite{liuo0csmskc24} adopts data-free distillation, which generates synthetic data from pretrained models to guide the 4-bit quantization of LLaMA architectures without the need for original training data. BitDistiller~\cite{duzcgccx24} leverages full-precision models as teachers and introduces a tailored asymmetric quantization and clipping strategy to minimize error in sub-4-bit regimes. 

\textbf{Other QAT methods.}
Additionally, several methods push the limits of quantization techniques. EfficientQAT~\cite{chen2024efficientqat} addresses the memory challenges associated with traditional QAT methods. It employs a two-stage training strategy. It first trains the model and quantization parameters block-by-block, followed by end-to-end training of the quantization parameters alone. QuEST~\cite{panferov2025quest} improves both accuracy and speed by introducing Hadamard normalization and MSE-optimal fitting, alongside a novel gradient estimator called trust estimation, which ensures stable training at 1-bit precision. Theoretical insights from \mcite{Pang}{pang2025stabilizing} reveal that QAT instability arises from loss landscape sharpness. To address this, they introduce Feature-Perturbed Quantization (FPQ), which smooths the loss landscape through implicit Hessian regularization.

%% file: fig/forest_qat.tex
\tikzstyle{leaf}=[
    align=left,
    inner xsep=10pt,
    inner ysep=3pt,
]

\begin{figure}[t]
    \centering
    \begin{forest}
        forked edges,
        for tree={
            font=\sffamily\bfseries, 
            text=textcolor, 
            grow=east,
            reversed=true,
            anchor=base west,
            parent anchor=east,
            child anchor=west,
            base=center,
            rectangle,
            draw=black,
            rounded corners,
            align=left,
            text centered,
            minimum width=4em,
            edge+={black, line width=1pt},
            s sep=3pt,
            inner xsep=2pt,
            inner ysep=3pt,
            line width=0.8pt,
            ver/.style={rotate=90, child anchor=north, parent anchor=south, anchor=center},
            myline/.style={line width=0.8pt},
        },
        where level=0{fill=rootcolor,text=white,text width=6em,font=\footnotesize,}{},
        where level=1{fill=level3color,text width=18em,font=\scriptsize,leaf,}{},
        [
        {QAT Techniques}, ver,
                [
                    {
                    \textbf{Binary and ternary QAT:} BitNet~\cite{wang2023bitnet},
                    BitNet b1.58~\cite{ma2024era},\\
                    \mcite{Nielsen}{nielsen2025continual}
                    }
                    ,
                ]
                [
                    {
                    \textbf{QAT with KD:} LLM-QAT~\cite{liuo0csmskc24},
                    BitDistiller~\cite{duzcgccx24}
                    }
                    ,
                ]
                [
                    {
                    \textbf{Other QAT methods:} EfficientQAT~\cite{chen2024efficientqat},
                    QuEST~\cite{panferov2025quest},\\
                    \mcite{Pang}{pang2025stabilizing}
                    }
                    ,
                ]
        ]
    \end{forest}
    \caption{Overview of studies incorporating quantization-aware training techniques.}
    \label{fig:qat}
\end{figure}